\documentclass[11pt]{article}
\usepackage[utf8]{inputenc}

\usepackage[preprint]{acl}
\usepackage{tabularx} 

\usepackage[table]{xcolor}
\usepackage{times}
\usepackage{latexsym}
\usepackage{amssymb}
\usepackage{algorithm}
\usepackage{graphicx}
\usepackage{subcaption}
\usepackage{algpseudocode}
\usepackage{hyperref}
\usepackage[T1]{fontenc}
\usepackage{amsmath}
\usepackage{booktabs}

\usepackage[utf8]{inputenc}

\usepackage{microtype}

\usepackage{inconsolata}

\title{ReasonAny: Incorporating Reasoning Capability to Any Model via Simple and Effective Model Merging}

\author{
Junyao Yang \textsuperscript{\rm 1,2},
Chen Qian \textsuperscript{\rm 1,3}\thanks{Project lead. This work was done during Junyao Yang's internship at Shanghai Artificial Intelligence Laboratory, supervised by Chen Qian.}, 
Dongrui Liu \textsuperscript{\rm 1}\textsuperscript{\dag}, 
Wen Shen \textsuperscript{\rm 4},
Yong Liu \textsuperscript{\rm 3}\textsuperscript{\dag}, 
Jing Shao \textsuperscript{\rm 1}\textsuperscript{\dag}\\ 
\textsuperscript{\rm 1} Shanghai Artificial Intelligence Laboratory\\
\textsuperscript{\rm 2} National University of Singapore \
\textsuperscript{\rm 3} Renmin University of China \
\textsuperscript{\rm 4} Tongji University\\
{$\;\:$ \texttt{junyaoyang@u.nus.edu}} {$\;\:$ \texttt{\{qianchen2022,liuyonggsai\}@ruc.edu.cn}} \\
{$\;\:$ \texttt{wenshen@tongji.edu.cn}} {$\;\:$ \texttt{\{liudongrui, shaojing\}@pjlab.org.cn}}
}

\begin{document}
\maketitle
\begingroup
\renewcommand\thefootnote{\dag}
\footnotetext{Corresponding author.}
\endgroup


\begin{abstract}
Large Reasoning Models (LRMs) with long chain-of-thought reasoning have recently achieved remarkable success.
Yet, equipping domain-specialized models with such reasoning capabilities, referred to as ``Reasoning + X'', remains a significant challenge. 
While model merging offers a promising training-free solution, existing methods often suffer from a {destructive performance collapse: existing methods tend to both weaken reasoning depth and compromise domain-specific utility.}
Interestingly, we identify a counter-intuitive phenomenon underlying this failure: \textit{reasoning ability predominantly resides in parameter regions with low gradient sensitivity, contrary to the common assumption that domain capabilities correspond to high-magnitude parameters}.
Motivated by this insight, we propose \textbf{ReasonAny}, a novel merging framework that resolves the {reasoning–domain performance collapse} through Contrastive Gradient Identification.
Experiments across safety, biomedicine, and finance domains show that ReasonAny effectively synthesizes ``Reasoning + X'' capabilities, significantly outperforming state-of-the-art baselines while retaining robust reasoning performance. Code is available at \href{https://github.com/jyyang26/ReasonAny}{Github}.
\end{abstract}

\section{Introduction}
\begin{figure}[t!]
    \centering
    \includegraphics[width=0.95\linewidth]{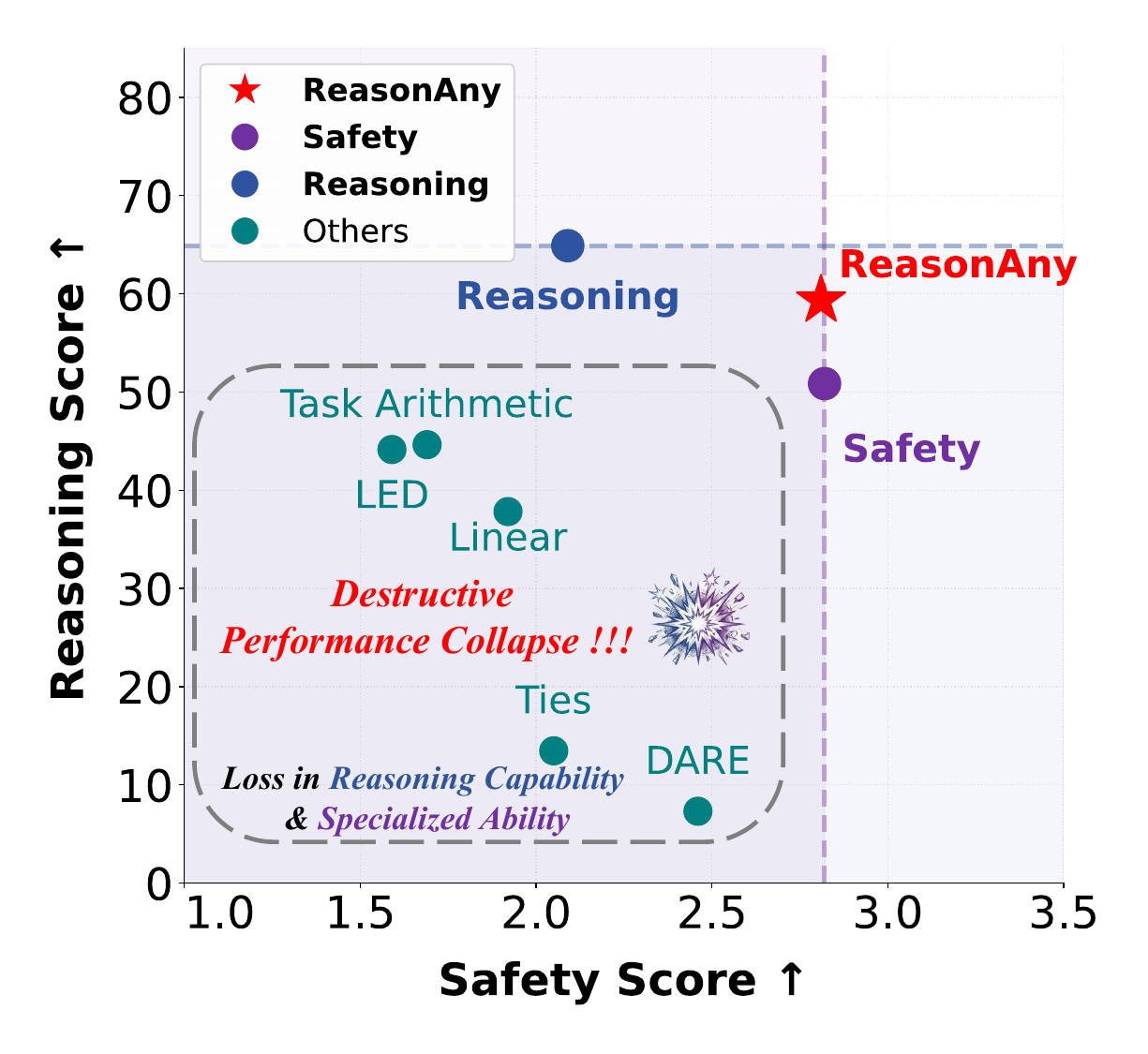}
    \caption{ReasonAny overcomes the {destructive performance collapse} in model merging, evaluated via GSM8K accuracy and \textit{max - current harmfulness score} as Safety Score on Safety-Tuned bench. Methods in purple and blue bounds show the loss in specialized ability and reasoning capability, respectively. By reaching the top-right corner, ReasonAny preserves robust reasoning capability without compromising specialized utility.}
    \label{fig:intro_fig}
    \vspace{-1em}
\end{figure}
\begin{figure*}[ht!]
\centering
\includegraphics[width=0.98\textwidth]{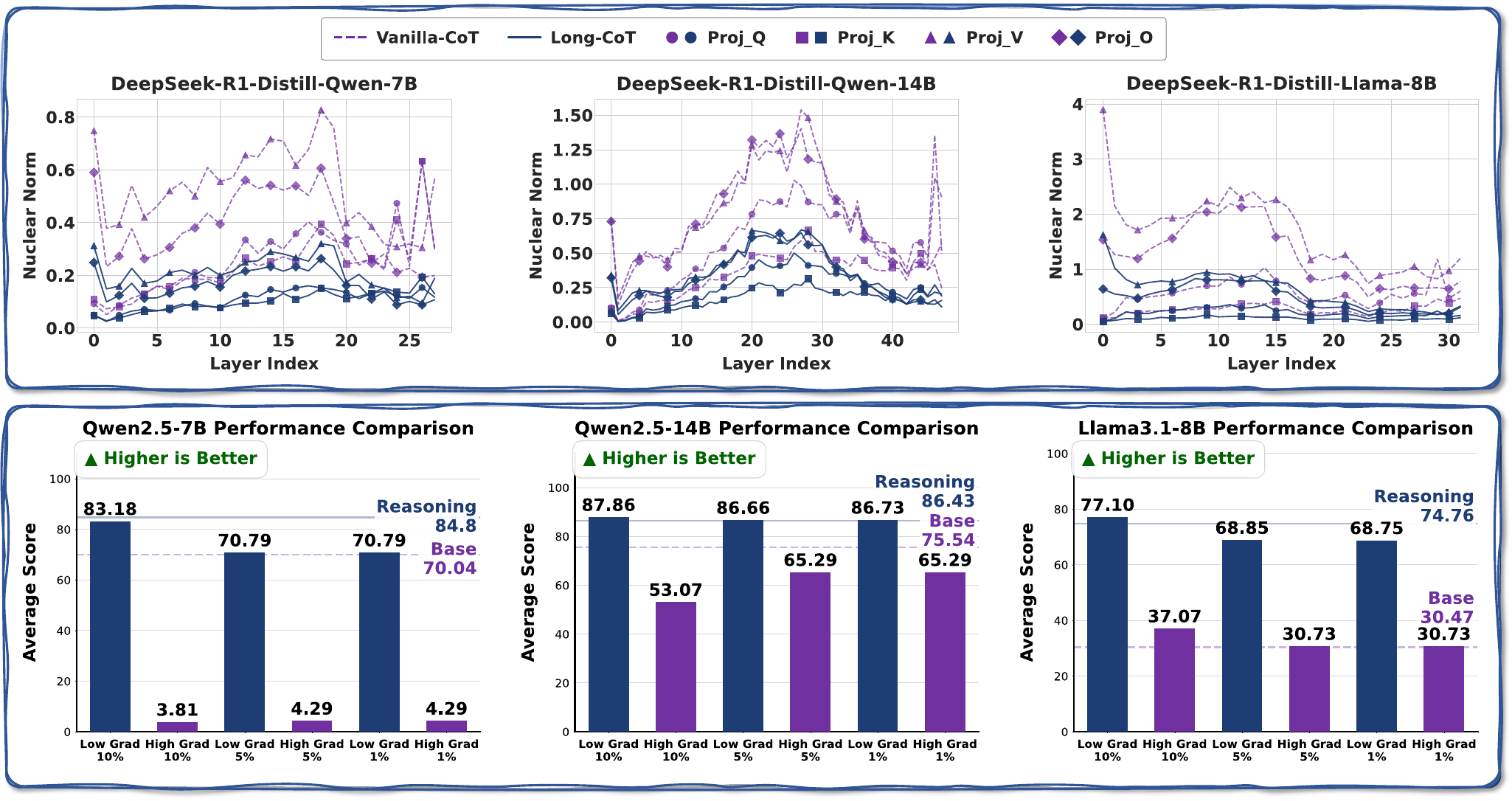}
\vspace{-0.3em}
\caption{\textbf{Gradients Nuclear Norm Analysis and Additive Experiment Results.} The \textbf{top sub-figure} shows gradient analysis across ($Q, K, V, O$) projection matrices at all layers. The \textbf{top-left}, \textbf{top-middle}, and \textbf{top-right} panels display Nuclear Norms for DeepSeek-R1-Distill Qwen-7B, Qwen-14B, and Llama-8B respectively, revealing that long-CoT induces significantly lower gradients than Short-CoT. The \textbf{bottom sub-figures} display additive experiments validating that reasoning capability lies in low-gradient regions. By merging weights from 10\%, 5\%, and 1\% of highest and lowest gradient into base models, results across the \textbf{top-left}, \textbf{top-middle}, and \textbf{top-right} sub-figures consistently demonstrate that reasoning capability depends on weights associated with low gradients.}
\label{fig:gradient_analysis}
\vspace{-0.7em}
\end{figure*}

The recent emergence of Large Reasoning Models (LRMs) represents a milestone breakthrough in the landscape of Large Language Models (LLMs) \citep{grattafiori2024llama,qwen2.5}. 
By leveraging the long chain-of-thought (long-CoT) mechanisms \citep{yeotong2025longcot}, reasoning models have demonstrated exceptional performance, particularly in specialized tasks such as mathematics and coding \citep{jaech2024openai,qwq32b,guo2025deepseek,openai2025o3}. 
Still, equipping models in specific domain tasks with these advanced reasoning capabilities is a vital yet under-explored frontier. 
For LLMs equipped with domain-specific knowledge, such as safety alignment \citep{kuo2025hcot}, biomedicine \citep{Ullah2024biolack, Griot2025biolack}, or finance \citep{Zhao2024financellm,Yuqi2024financesurvey}, one objective is to construct models that have not only robust \textbf{Reasoning} capabilities but are also specialized in domain-specific tasks ``\textbf{X}''.
We term this critical synthesis ``\textbf{Reasoning + X}''.

To achieve this synthesis, the prevailing approach involves Supervised Fine-Tuning (SFT) or Reinforcement Learning (RL) on domain-specific reasoning datasets \citep{kuo2025hcot,qian2025fino1transferabilityreasoningenhanced,p12025,Kai-tao2025_med_reason,chen-etal-2025-towards-medical_reason,Bao2025SafeR1CS}. 
Despite its efficacy, this paradigm faces challenges: difficulty constructing domain-specific reasoning data \citep{Chen2024HuatuoGPTo1TM,qian2025fino1transferabilityreasoningenhanced}, resource-intensive training \citep{Matsutani2025}, and catastrophic forgetting \citep{PARISI201954,parthasarathy2024ultimate}.
In light of these challenges, model merging has emerged as a compelling, training-free alternative designed to combine distinct capabilities from different models into a single entity \citep{yang2024modelmergesurvey,ZHOU2025100102survey2,lan2025thinking}.

Motivated by this potential, 
we conduct a preliminary exploration to merge reasoning and domain-specific models via state-of-the-art techniques.
Interestingly, as illustrated in Figure \ref{fig:intro_fig}, our experiments reveal a \textbf{Destructive Performance Collapse} in the context of reasoning—resulting merged models typically suffer from both significant loss in reasoning capability and severe compromise in the specialized abilities of ``X''.
This phenomenon persists despite existing methods \citep{yadav2023tiesmerging,liu2025sens} proving effective for standard knowledge injection.
Such a setback likely stems 
from the common assumption that high-magnitude weights or gradients identify important parameters  \citep{yadav2023tiesmerging,hao2025training}. 
Our findings challenge this intuition and raise a pivotal question:
\textbf{Do parameters handling reasoning capability follow the same high-magnitude rules as knowledge locating?}


As illustrated in the top part of Figure \ref{fig:gradient_analysis}, we uncover a \textbf{counter-intuitive phenomenon:} \textbf{\textit{reasoning capabilities are characterized by subtle, low-magnitude gradient changes, challenging the prevailing belief that important features necessarily generate high-magnitude gradients shifts }}\citep{,liu2025sens,ma2025led}.
This phenomenon reveals that models employing long-CoT and models with superior reasoning capabilities consistently exhibit significantly lower gradient magnitudes compared to standard instruct-tuned models.

\begin{figure*}[t]
\centering
\includegraphics[width=0.98\textwidth]{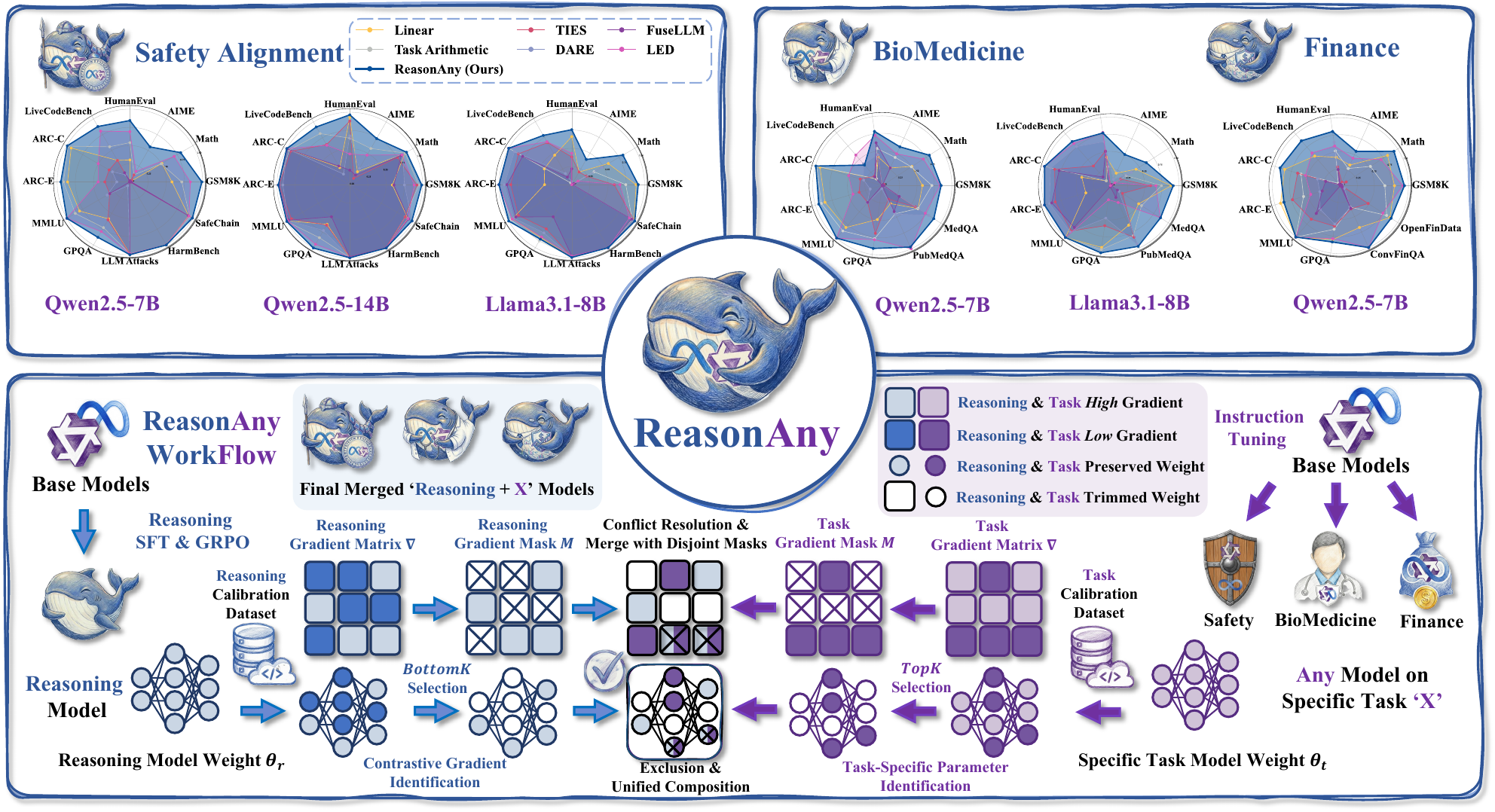}
\vspace{-0.3em}
\caption{\textbf{Experimental Results and Workflow of ReasonAny}. Experimental results on \textbf{Safety} (\textbf{top-left}), \textbf{Biomedicine}, and \textbf{Finance} (\textbf{top-right}) benchmarks demonstrate ReasonAny, shown in light blue background, significantly outperforming baselines. \textbf{ReasonAny {Workflow}} (\textbf{bottom}) employs \textbf{Contrastive Gradient Identification} (\textbf{bottom-right}) to isolate low-gradient reasoning and high-gradient task weights and \textbf{Exclusion} (\textbf{bottom-middle}) step disjoint masks that merge specialized capabilities without compromising reasoning capabilities.}
\label{fig:main_algorithm}
\vspace{-0.7em}
\end{figure*}

Based on this counter-intuitive phenomenon, we propose \textbf{ReasonAny}, a novel merging framework designed to resolve the ``Reasoning + X'' conflict. 
Unlike traditional methods that treat all tasks uniformly under a single importance metric \citep{zeng25revisiting,Thapa2025DisentanglingRA}, ReasonAny employs \textbf{Contrastive Gradient Identification} to handle these conflicting model parameters selection. 
Specifically, we isolate the robust features of domain-specific task ``X'' using traditional high-gradient selection, while simultaneously capturing reasoning capabilities through a targeted \textit{low-gradient} filtering mechanism. 
To ensure these distinct capabilities coexist without destructive performance collapse, we implement \textbf{Conflict Resolution via Exclusion} that creates mutually exclusive parameter masks before composing the final model. The overall workflow of ReasonAny is shown in the bottom part of Figure \ref{fig:main_algorithm}. 
Our experiments demonstrate that ReasonAny successfully incorporates advanced reasoning capabilities into diverse models without compromising their domain-specific capabilities, offering a simple yet effective solution to the reasoning-utility performance collapse.

\section{Reasoning Capabilities Reside in Low-Gradient Parameter Regions}
\label{sec:low_gradients}

In the pursuit of synthesizing reasoning capabilities with domain-specific task ``X'', model merging presents a promising training-free solution \citep{ilharco2023editing,yang2024modelmergesurvey,ZHOU2025100102survey2}. 
However, as illustrated in Figure \ref{fig:intro_fig} and comprehensive evaluation in Section \ref{sec:main_exp}, we observe that traditional merging methods often suffer from \textbf{Destructive Performance Collapse} with both the reasoning capability collapses and the domain utility is compromised.  

To resolve this, we investigate the distinct gradient characteristics underlying reasoning capabilities. 
Specifically, Section \ref{sec:preliminaries} establishes the mathematical foundations for model merging, Section \ref{subsec:nuclear_norm_analysis} analyzes the unique gradient magnitude distributions of reasoning models, and Section \ref{subsec:add_gradients} confirms that reasoning specifically relies on low-gradient structures through targeted additive experiments.

\subsection{Preliminaries}
\label{sec:preliminaries}

\paragraph{Models and Task Vectors.}
We operate within the parameter space of Transformer-based LLMs. Let $\theta_{\text{base}} \in \mathbb{R}^d$ denote the parameters of a pre-trained base model. We consider a scenario where $\theta_{\text{base}}$ serves as the initialization for two distinct fine-tuning processes:
(1) A \textbf{Task Model} {$\theta_t \in \mathbb{R}^d$}, which is fine-tuned on a domain-specific task ``X'' dataset $\mathcal{D}_t$, such as safety, biomedicine or finance.
(2) A \textbf{Reasoning Model} $\theta_r \in \mathbb{R}^d$, which is fine-tuned on a reasoning-intensive dataset $\mathcal{D}_r$. 

Following standard arithmetic merging formulations \citep{ilharco2023editing}, we define the \textit{task vector} $\tau$ as the dense displacement in the parameter space resulting from fine-tuning. The task vectors for the specialized task and reasoning are defined respectively as:
\vspace{-0.5em}
\begin{equation}
\tau_t = \theta_t - \theta_{\text{base}}, \quad
\tau_r = \theta_r - \theta_{\text{base}}.
\end{equation}

\vspace{-0.5em}

Intuitively, these vectors encode the specific model weight shifts required to endow the base model with specialized domain expertise or reasoning capabilities. 
Our objective is to construct a merged parameter set $\theta_{\text{merged}}$ that incorporates the functional capabilities of both $\tau_t$ and $\tau_r$ without destructive interference.

\paragraph{Gradient-Based Parameter Identification.}
To determine the topology of the critical parameters for each task, prevailing methodologies in recent works extensively utilize gradient-based metrics for parameter identification \citep{liu2025sens,ma2025led}.
For a given model parameterized by $\theta$ and a calibration dataset $\mathcal{D}$, the importance score $I_j$ for the $j$-th parameter is computed as the expectation of the gradient magnitude with respect to the loss function $\mathcal{L}$. 
Formally, the identification vector $I(\theta, \mathcal{D}) \in \mathbb{R}^d$ is defined as:
\begin{equation}
I(\theta, \mathcal{D}) = \mathbb{E}_{x \sim \mathcal{D}} \left[ \left| \nabla_{\theta} \mathcal{L}(x; \theta) \right| \right].
\end{equation}

\vspace{-0.4em}

This metric proxies parameter saliency, quantifying task performance sensitivity to weight perturbations.
Intuitively, higher values indicate task-critical weights.


\subsection{Gradient Magnitude Distribution Analysis}
\label{subsec:nuclear_norm_analysis}
Prevailing research generally operates on the assumption that high-magnitude gradients encode the most critical model capabilities \citep{liu2025sens,ma2025led}. 
Adopting the spectral analysis from \citet{li-etal-2025-happened}, 
we measure gradient magnitude distribution across layers for both Task Model and Reasoning Model using \textbf{Nuclear Norm}:
\begin{equation}
    s_{x,i} = \|\nabla_{i} \mathcal{L}(x; i)\|_* = \sum_{j=1}^{\min\{m,n\}} \sigma_j,
\end{equation}
where $\sigma_j$ represents the singular values of the gradient matrix $\nabla_{i} \mathcal{L}(x; i)$ 
corresponding to the $Q, K, V,$ and $O$ projection matrices at layer $i$.

As shown in top sub-figures of Figure \ref{fig:gradient_analysis}, Qwen2.5-7B, Qwen2.5-14B and Llama3.1-8B series reasoning models with blue line marked as Long-CoT, exhibit significantly \textbf{lower nuclear norms} than the base model purple line marked as Vanilla-CoT. 
This \textbf{counter-intuitive phenomenon} suggests that reasoning capabilities reside in low-gradient regions, challenging conventional assumptions that high-magnitude gradients corresponding weights encode more important information.

Detailed analysis of correlation between gradients and nuclear norms is shown in Appendix \ref{app:gradients_nuclear_norm_mad}.



\subsection{How Do Low-gradient Parameters Work?}
\label{subsec:add_gradients}
To empirically validate whether reasoning capabilities are localized within low-gradient model weights, we conduct an additive experiment based on the intuition that exclusively injecting these targeted weights into the base model should significantly reactivate its reasoning abilities.
Specifically, we selectively add parameters from the reasoning task vector $\tau_r$ into the base model $\theta_{\text{base}}$ via:

\vspace{-0.2em}
\begin{equation}
\theta' = \theta_{\text{base}} + \tau_r \odot \mathbf{M},
\end{equation}
\vspace{-0.7em}

\noindent where $\mathbf{M}$ denotes a binary mask that filters weights based on their gradient magnitude ranking.

We apply this method to base model by adding the \textit{Highest Gradients }and the \textit{Lowest Gradients} corresponding parameters at specific sparsity ratios of 10\%, 5\% and 1\%. 
As illustrated in the bottom of Figure \ref{fig:gradient_analysis}, Qwen2.5-7B recovers GSM8K scores of $83.18$ and $70.79$ using 10\% and lower ratios of lowest gradient parameters, whereas incorporating highest gradient updates leads to a complete performance collapse.
This phenomenon can also be found on both Qwen2.5-14B and Llama3.1-8B series models.
\textbf{\textit{This distinct performance disparity strongly validates that reasoning capabilities are predominantly localized within low-magnitude gradient model weight regions.}}

\section{Methodology}
\label{sec:methodology}

\setlength{\textfloatsep}{10pt}
\begin{algorithm}[t]
\caption{\textsc{ReasonAny}}
\label{alg:reasonany}
\begin{algorithmic}[1]
\Require Base model $\theta_{\text{base}}$, Task model $\theta_t$, Reasoning model $\theta_r$, Calibration datasets $\mathcal{D}_t, \mathcal{D}_r$, Selection ratios $p_t, p_r$, Scaling factors $\lambda_t, \lambda_r$
\Ensure Merged parameters $\theta_{\text{merged}}$

\State \textbf{Initialize} $\theta_{\text{merged}} \leftarrow \theta_{\text{base}}$

\State \textit{// Step 1: Calculate Task Vectors}
\State $\tau_t \leftarrow \theta_t - \theta_{\text{base}}$, $\tau_r \leftarrow \theta_r - \theta_{\text{base}}$

\State \textit{// Step 2: Calculate Importance Scores (Gradient Sensitivity)}
\State $I(\theta_t) \leftarrow \mathbb{E}_{x \sim \mathcal{D}_t} [ |\nabla_{\theta} \mathcal{L}(x; \theta_t)| ]$
\State $I(\theta_r) \leftarrow \mathbb{E}_{x \sim \mathcal{D}_r} [ |\nabla_{\theta} \mathcal{L}(x; \theta_r)| ]$

\State \textit{// Step 3: Identify Subspaces}
\State $d \leftarrow \text{length}(\theta_{\text{base}})$
\State {$\mathcal{N}_t \leftarrow \text{TopK}(I(\theta_t), p_t)$} \Comment{High-gradient for Task}
\State {$\mathcal{N}_r \leftarrow \text{BottomK}(I(\theta_r), p_r)$ }\Comment{Low-gradient for Reasoning}

\State \textit{// Step 4: Conflict Resolution (Exclusion)}
\State $\mathcal{T}'_t \leftarrow \mathcal{N}_t \setminus \mathcal{N}_r, \quad \mathcal{T}'_r \leftarrow \mathcal{N}_r \setminus \mathcal{N}_t$

\State \textit{// Step 5: Merge with Disjoint Masks}
\State \textbf{Initialize Masks} $\mathbf{M}_t \leftarrow \mathbf{0}, \mathbf{M}_r \leftarrow \mathbf{0}$
\For{$i \in \mathcal{T}'_t$} $\mathbf{M}_{t,i} \leftarrow 1$ \EndFor
\For{$j \in \mathcal{T}'_r$} $\mathbf{M}_{r,j} \leftarrow 1$ \EndFor

\State $\theta_{\text{merged}} \leftarrow \theta_{\text{merged}} + \lambda_t (\tau_t \odot \mathbf{M}_t) + \lambda_r (\tau_r \odot \mathbf{M}_r)$

\State \Return $\theta_{\text{merged}}$
\end{algorithmic}
\end{algorithm}

\subsection{Overview of {ReasonAny}}
We introduce {ReasonAny}, a unified framework designed to synthesize the capabilities of a generic specialized \textbf{Task Model} (e.g., Safety, Biomedicine, Finance) and a \textbf{Reasoning Model} into a single backbone. 
ReasonAny operate this pipeline through two distinct stages: we first employ \textbf{Contrastive Gradient Identification} to isolate capability-specific parameter regions. 
Subsequently, we separate and culminate via certain model weights \textbf{Exclusion and Unified Composition} to synthesize these disjoint sets without destructive interference. 
The workflow and algorithm are illustrated in Figure \ref{fig:main_algorithm} and Algorithm \ref{alg:reasonany}.

\subsection{Parameter Identification}
\paragraph{Reasoning Parameter Identification.}
Recalling the insight that reasoning capabilities are encoded in reasoning model weights exhibiting low-magnitude gradients, we adopt a \textbf{Contrastive Gradient Identification} strategy in the first phase of ReasonAny. We select model weights with the {lowest} gradient magnitude on the reasoning dataset $\mathcal{D}_r$ and let $\text{BottomK}(v, k)$ be an operator returning the smallest values ratio $k$ in {task vector $v$}. The elected reasoning weight set $\mathcal{N}_r$ is defined as:

\vspace{-0.5em}

\begin{equation}
\mathcal{N}_r = \text{BottomK}\big(I(\theta_r, \mathcal{D}_r), p_r ),
\end{equation}
where $p_r$ represents the selection ratio for total reasoning model parameters $\theta_r$.

\paragraph{Task-Specific Parameter Identification.}
In parallel, we identify the parameters critical for the specialized Task ``X'' such as safety alignment, biomedicine and finance expertise. 
Consistent with established pruning and merging literature \citep{liu2025sens,ma2025led,yang2025rcp}, with results shown in Section \ref{subsec:add_gradients}, we reaffirm that domain-specific knowledge is retained in parameters with high sensitivity to the task loss. 
Therefore, we employ a standard \textit{Top-K} selection strategy on the task model $\theta_t$ using dataset $\mathcal{D}_t$. Let $\text{TopK}(v, k)$ denote the largest values indices with the ratio $k$. The elected task parameter set $\mathcal{N}_t$ is:
\begin{equation}
\mathcal{N}_t = \text{TopK}\big(I(\theta_t, \mathcal{D}_t), p_t),
\end{equation}
where $p_t$ is the selection ratio for the task model $\theta_t$.

\subsection{Exclusion and Unified Composition}

\paragraph{Conflict Resolution via Exclusion.}
A fundamental challenge in merging distinct models is parameter conflict, where a single parameter is deemed critical for both reasoning and the specific task ($\mathcal{N}_r \cap \mathcal{N}_t \neq \emptyset$). To preventing destructive interference—where the injection of domain knowledge might degrade reasoning depth—we enforce mutual exclusivity through a set-theoretic exclusion process. We derive the final, disjoint parameter sets $\mathcal{T}'_r$ and $\mathcal{T}'_t$ by removing overlapping indices, ensuring that each parameter is updated by at most one source using
$\mathcal{T}'_r = \mathcal{N}_r \setminus \mathcal{N}_t$ and $ \mathcal{T}'_t = \mathcal{N}_t \setminus \mathcal{N}_r. $
This step guarantees that the delicate low-gradient structures preserved for reasoning are not overwritten by high-magnitude task updates.

\paragraph{Unified Model Composition.}
Finally, we construct the unified model by composing the base model with the disjointly selected task vectors. We define binary masks $\mathbf{M}_r, \mathbf{M}_t \in \{0, 1\}^d$ corresponding to the indices in $\mathcal{T}'_r$ and $\mathcal{T}'_t$ respectively. 
The final merged weights $\theta_{\text{merged}}$ are computed as:
\begin{equation}
\theta_{\text{merged}} = \theta_{\text{base}} + \lambda_r (\tau_r \odot \mathbf{M}_r) + \lambda_t (\tau_t \odot \mathbf{M}_t),
\end{equation}
where $\odot$ denotes the element-wise product, and $\lambda_r, \lambda_t$ are scaling factors. This formulation effectively merges the ``Reasoning + X'' capabilities into the base model while strictly respecting the topological boundaries identified in the previous steps.

\section{Experiments}
\label{sec:main_exp}
\begin{table*}[t!]
\centering
\caption{Performance comparison of merging Qwen2.5-7B family with safety fine-tuning Qwen2.5-7B-Instruct (Safety) and DeepSeek-R1-Distill-Qwen-7B (Reasoning) on all datasets across Reasoning, Knowledge and Safety Benchmarks, where \textbf{Average} $\uparrow$ column indicate average performance across performance bench. The best performance among all merging methods on each dataset is highlighted in \textbf{bold}, and values highlighted in \textit{italic} with $^*$ mark indicate model output collapse.}
\label{tab:qwen_7b_safety_performance}
\renewcommand{\arraystretch}{1.1} 
\resizebox{\textwidth}{!}{
\begin{tabular}{l|ccccccccc|c|ccc}

\hline

{\textbf{Eval Bench}} & \multicolumn{10}{c|}{\textbf{Performance Bench}} & \multicolumn{3}{c}{\textbf{Safety Bench}} \\

\cline{1-14}

\textbf{Sub Areas}& \multicolumn{5}{c|}{\textbf{Reasoning}} & \multicolumn{4}{c|}{\textbf{Knowledge}}& \multicolumn{1}{c|}{\textbf{}} & \multicolumn{3}{c}{\textbf{Safety}} \\

\cline{1-14}

\textbf{Datasets} & \textbf{GSM8K}$\uparrow$ & \textbf{Math}$\uparrow$ & \textbf{AIME}$\uparrow$ & \textbf{HumanEval}$\uparrow$ & \textbf{LiveCodeBench}$\uparrow$ & \textbf{ARC-C}$\uparrow$ & \textbf{ARC-E}$\uparrow$ & \textbf{MMLU}$\uparrow$ & \textbf{GPQA}$\uparrow$ & \textbf{Average}$\uparrow$ & \textbf{Safety-Tuned}$\downarrow$ & \textbf{HarmBench}$\downarrow$ & \textbf{SafeChain}$\uparrow$ \\

\hline
\textbf{Safety} & 69.42 & 74.00 & 13.33 & 50.64 & 12.43 & 60.91 & 65.78 & \textbf{71.72} & 39.39 & 50.85 & \textbf{1.18} & \textbf{0.08} & \textbf{4.90} \\
\textbf{Reasoning} & \textbf{87.23} & \textbf{86.20} & \textbf{60.00} & \textbf{76.63} & \textbf{30.37} & \textbf{64.75} & \textbf{77.25} & 52.51 & \textbf{49.10} & \textbf{64.89} & 1.91 & 0.46 & 4.61 \\
\hline
\textbf{Linear} & 50.42 & 43.80 & 0.00 & 23.14 & 10.38 & 60.34 & 66.19 & 57.34 & 28.78 & 37.82 & {2.08} & 0.31 & 4.57 \\
\textbf{Task Arithmetic} & 62.17 & 42.80 & 6.67 & 41.35 & 16.93 & 63.56 & 70.43 & 62.14 & 35.61 & 44.63 & {2.31} & {0.40} & {4.66} \\
\textbf{Ties} & \textit{0.83}$^*$ & \textit{2.60}$^*$ & \textit{6.67}$^*$ & \textit{0.00}$^*$ & \textit{10.38}$^*$ & {21.36} & \textit{26.63} & \textit{23.08} & \textit{29.55}$^*$ & \textit{13.46}$^*$ & 1.95 & \textit{0.02}$^*$ & {4.86} \\
\textbf{DARE} & \textit{0.53}$^*$ & \textit{1.00}$^*$ & \textit{0.00}$^*$ & \textit{0.00}$^*$ & \textit{3.38}$^*$ & {17.97} & {13.93} & {25.95} & \textit{3.03}$^*$ & \textit{7.31}$^*$ & 1.54 & \textit{0.00}$^*$ & {4.86} \\
\textbf{FuseLLM} & \textit{1.81}$^*$ & \textit{0.20}$^*$ & \textit{0.00}$^*$ & \textit{1.23}$^*$ & \textit{4.43}$^*$ & \textit{0.00}$^*$ & \textit{0.00}$^*$ & {22.95} & \textit{0.00}$^*$ & \textit{3.40}$^*$ & \textbf{0.87} & \textit{0.01}$^*$ & 4.54 \\
\textbf{LED} & 72.48 & 60.60 & 10.00 & 52.91 & 24.51 & 32.54 & 33.69 & 71.93 & 38.64 & 44.14 & {2.41} & {0.36} & 4.59 \\
\rowcolor{lightgray!25} \textbf{ReasonAny} & \textbf{86.28} & \textbf{69.40} & \textbf{33.33} & \textbf{64.65} & \textbf{26.71} & \textbf{64.31} & \textbf{73.39} & \textbf{72.73} & \textbf{43.18} & \textbf{59.33} & 1.19 & \textbf{0.08} & \textbf{4.86} \\
\hline
\end{tabular}}
\vspace{-0.7em}
\end{table*}

\subsection{Experiments Setup}
\label{subsec:setup}
\paragraph{Baselines.}
We compared ReasonAny with multiple merging baselines: \textbf{Linear} \citep{Izmailov2018Averaging}, \textbf{Task Arithmetic} \citep{ilharco2023editing}, \textbf{TIES-Merging} \citep{yadav2023tiesmerging}, \textbf{DARE-Merging} \citep{yu2024dare}, \textbf{FuseLLM} \citep{wan24fusellm} and \textbf{LED-Merging} \citep{ma2025led}. We utilize mergekit \citep{goddard-etal-2024-arcees-mergekit} as merging tools for baseline methods. Detailed baselines explanation and recommended hyperparameter settings are listed in Appendix \ref{app:baselines_explanation} and \ref{app:hyper_setting}. Moreover, explanation of Figure \ref{fig:intro_fig} is shown in Appendix \ref{sec:exp_fig1_setting}.

\paragraph{Datasets.}
Using the same \textbf{Performance} benchmark for \textbf{Reasoning} and \textbf{Knowledge}, we evaluated \textbf{Safety}, \textbf{Biomedicine}, and \textbf{Finance} tasks using domain-specific benchmarks. 
In performance benchmarks, for \textit{Reasoning Evaluation}, we assess with \textit{GSM8K} \citep{Cobbe21gsm}, \textit{Math500} \citep{lightman2023letsmath500} and \textit{AIME2024} \citep{aime_1983_2024} for math reasoning, \textit{HumanEval} \citep{Mark21humaneval} and \textit{LiveCodeBench} \citep{jain2024livecodebench} for code reasoning. For \textit{Knowledge Evaluation}, we utilized \textit{ARC-E}, \textit{ARC-C} \citep{Clark18arc}, \textit{MMLU} \citep{hendryckstest2021mmlu1,hendrycks2021ethicsmmlu2} and \textit{GPQA} \citep{rein2023gpqa} to test the knowledge preservation of merged models. For \textit{Safety Evaluation}, \textit{Safety-Tuned} \citep{bianchi2024safetytuned}, \textit{HarmBench} \citep{mazeika2024harmbench} and \textit{SafeChain} \citep{jiang2025safechain} are used to verified the robustness of merged models. For \textit{BioMedicine Evaluation}, we use \textit{PubMedQA} \citep{Jin19PubMedQA} and \textit{MedQA} \citep{Jin20MedQA}. For \textit{Finance Evaluation}, we use \textit{ConvFinQA} \citep{chen2022convfinqa} and \textit{OpenFinData} \citep{openfindata}. 
We use opencompass \citep{2023opencompass} as the evaluation tool. 
Detailed datasets explanation is shown in Appendix \ref{app:datasets_explanation}.

\paragraph{Models.}
Our experiments utilize base models on the \textbf{Qwen2.5} and \textbf{Llama-3.1} series \citep{qwen2.5, grattafiori2024llama}.
The corresponding reasoning models are \textbf{DeepSeek-R1-Distill} series models and \textbf{QwQ-32B-Preview} \citep{guo2025deepseek,qwq32b}. 
For safety task, by fine-tuning on Safety training Dataset \citep{bianchi2024safetytuned} using Low-Rank Adaptation \citep{hu2022lora,wang2023alpacalora} on corresponding instruct models, we obtain the model with the best safety performance among the corresponding family of models in our setting.
For biomedicine task, we use \textbf{Meditron3-Qwen2.5-7B} and \textbf{MMed-Llama-3-8B} on Qwen2.5-7B and Llama3.1-8B family as biomedicine task expert \citep{Chen23Meditron,qiu2024building-mmed-llama}.
For finance task, we use \textbf{WiroAI-Finance-Qwen-7B} and \textbf{WiroAI-Finance-Llama-8B} on Qwen2.5-7B and Llama3.1-8B family as finance task expert \citep{WiroAI-finance-qwen-7b,WiroAI-finance-llama-8b}.
Full model configuration are shown in Appendix \ref{app:models_config}.





\begin{table*}[t!]
\centering
\caption{Performance comparison of merging Qwen2.5-7B family with Meditron3-Qwen2.5-7B (Biomedicine) and DeepSeek-R1-Distill-Qwen-7B (Reasoning) on all datasets across Reasoning, Knowledge and Biomedicine Benchmarks, where \textbf{Average} $\uparrow$ column indicate average performance across performance bench. The best performance among all merging methods on each dataset is highlighted in \textbf{bold}.}
\vspace{-0.2em}
\label{tab:qwen_7b_bio_performance}
\renewcommand{\arraystretch}{1.1} 
\resizebox{\textwidth}{!}{
\begin{tabular}{l|ccccccccc|cc|c}

\hline

{\textbf{Eval Bench}} & \multicolumn{9}{c|}{\textbf{Performance Bench}} & \multicolumn{2}{c|}{\textbf{Domain Bench}} \\

\cline{1-13}

\textbf{Sub Areas}& \multicolumn{5}{c|}{\textbf{Reasoning}} & \multicolumn{4}{c|}{\textbf{Knowledge}} & \multicolumn{2}{c|}{\textbf{Biomedicine}}& \multicolumn{1}{c}{\textbf{}} \\

\cline{1-13}

\textbf{Datasets} & \textbf{GSM8K}$\uparrow$ & \textbf{Math}$\uparrow$ & \textbf{AIME}$\uparrow$ & \textbf{HumanEval}$\uparrow$ & \textbf{LiveCodeBench}$\uparrow$ & \textbf{ARC-C}$\uparrow$ & \textbf{ARC-E}$\uparrow$ & \textbf{MMLU}$\uparrow$ & \textbf{GPQA}$\uparrow$ & \textbf{PubMedQA}$\uparrow$ & \textbf{MedQA}$\uparrow$&\textbf{Average}$\uparrow$ \\
\hline
\textbf{Biomedicine} & 69.40 & 74.00 & 6.67 & 37.95 & 3.20 & 60.34 & 67.02 & \textbf{71.51} & 40.15 & \textbf{51.00} & \textbf{54.46} & 48.70 \\
\textbf{Reasoning} & \textbf{87.23} & \textbf{86.20} & \textbf{60.00} & \textbf{89.61} & \textbf{30.37} & \textbf{64.75} & \textbf{77.25} & 52.51 & \textbf{49.10} & 38.00 & 30.20 & \textbf{60.47} \\
\hline
\textbf{Linear} & 50.42 & 43.80 & 16.67 & 37.23 & 3.80 & 60.34 & 66.19 & 57.04 & 24.24 & 14.60 & 33.36 & 37.06 \\
\textbf{Task Arithmetic} & 62.17 & 42.80 & 26.67 & 48.25 & 3.80 & 63.56 & 70.43 & 61.81 & 37.88 & 22.80 & 33.30 & 43.04 \\
\textbf{TIES} & 0.83 & 2.60 & 0.00 & 40.27 & 5.30 & 21.36 & 26.63 & 22.95 & 34.09 & 11.00 & 30.76 & 17.80 \\
\textbf{DARE} & 0.53 & 1.00 & 0.00 & 40.16 & 12.50 & 17.97 & 13.93 & 23.46 & 2.27 & 23.00 & 43.12 & 16.18\\
\textbf{FuseLLM} & 1.80 & 0.20 & 16.67 & 55.58 & 5.00 & 0.00 & 0.00 & 22.95 & 0.00 & 21.00 & 15.80 & 12.64 \\
\textbf{LED} & 72.48 & 60.60 & 30.00 & 65.23 & \textbf{17.60} & 32.54 & 33.89 & 71.93 & 38.64 & \textbf{56.40} & 40.06 & 47.20 \\
\rowcolor{lightgray!25} \textbf{ReasonAny} & \textbf{73.77} & \textbf{69.40} & \textbf{36.67} & \textbf{70.42} & 11.80 & \textbf{64.31} & \textbf{73.39} & \textbf{73.46} & \textbf{44.85} & 49.60 & \textbf{47.96} & \textbf{55.97} \\
\hline
\end{tabular}}
\vspace{-0.3em}
\end{table*}

\begin{table*}[t!]
\centering
\caption{Performance comparison of Qwen2.5-7B family ablation study when merging safety subbranch task model on Reasoning, Knowledge, and Safety Benchmarks, where \textbf{Average} $\uparrow$ column indicate average performance across performance bench. The best performance on each dataset is highlighted in \textbf{bold}.}
\vspace{-0.2em}
\label{tab:qwen_7b_ablation}
\renewcommand{\arraystretch}{1.2}
\resizebox{\textwidth}{!}{
\begin{tabular}{l|ccccccccc|c|ccc}

\hline

{\textbf{Eval Bench}} & \multicolumn{10}{c|}{\textbf{Performance Bench}} & \multicolumn{3}{c}{\textbf{Safety Bench}} \\

\cline{1-14}

\textbf{Sub Areas}& \multicolumn{5}{c|}{\textbf{Reasoning}} & \multicolumn{4}{c|}{\textbf{Knowledge}}& \multicolumn{1}{c|}{\textbf{}} & \multicolumn{3}{c}{\textbf{Safety}} \\

\cline{1-14}

\textbf{Datasets} & \textbf{GSM8K}$\uparrow$ & \textbf{Math}$\uparrow$ & \textbf{AIME}$\uparrow$ & \textbf{HumanEval}$\uparrow$ & \textbf{LiveCodeBench}$\uparrow$ & \textbf{ARC-C}$\uparrow$ & \textbf{ARC-E}$\uparrow$ & \textbf{MMLU}$\uparrow$ & \textbf{GPQA}$\uparrow$ & \textbf{Average}$\uparrow$ & \textbf{Safety-Tuned}$\downarrow$ & \textbf{HarmBench}$\downarrow$ & \textbf{SafeChain}$\uparrow$ \\

\hline
\textbf{Safety} & 69.42 & 74.00 & 13.33 & 50.64 & 12.43 & 60.91 & 65.78 & \textbf{71.72} & 39.39 & 50.85 & \textbf{1.18} & \textbf{0.08} & \textbf{4.90} \\
\textbf{Reasoning} & \textbf{87.23} & \textbf{86.20} & \textbf{60.00} & \textbf{76.63} & \textbf{30.37} & \textbf{64.75} & \textbf{77.25} & 52.51 & \textbf{49.10} & \textbf{64.89} & 1.91 & 0.46 & 4.61 \\
\hline
\textbf{w/o reason select} & 0.15 & 2.60 & 0.00 & 0.00 & 0.38 & 58.31 & 64.37 & 55.28 & 11.58 & 21.41 & 1.19 & 0.08 & 4.77 \\
\textbf{w/o safety select} & 86.13 & 76.00 & 16.67 & 31.32 & 7.00 & 63.05 & 65.78 & 71.71 & 42.24 & 51.10 & 2.39 & 0.18 & 4.56 \\
\rowcolor{lightgray!25} \textbf{ReasonAny} & \textbf{86.28} & \textbf{69.40} & \textbf{33.33} & \textbf{64.65} & \textbf{26.71} & \textbf{64.31} & \textbf{73.39} & \textbf{72.73} & \textbf{43.18} & \textbf{59.33} & \textbf{0.84} & \textbf{0.08} & \textbf{4.94} \\
\hline
\end{tabular}}
\vspace{-0.7em}
\end{table*}

\subsection{ReasonAny Preserves Specific Task Utility Alongside Robust Reasoning Capability}



\label{sec:reaonsing_safety_performance}



\paragraph{{ReasonAny ensures robust safety without compromising reasoning capability.}}
Table \ref{tab:qwen_7b_safety_performance} illustrates the performance comparing ReasonAny and baseline methods across Qwen2.5-7B benchmarks. 
ReasonAny retains a GSM8K score of $86.28$, recovering $98.91\%$ of reasoning capability. On the Safety Bench, ReasonAny adheres strictly to safety protocols. 
Conversely, 
Linear merging and DARE suffer catastrophic interference with GSM8K scores of $50.42$ and $0.53$, contrasting with the Reasoning expert's $87.23$. 
For the LLM Attacks benchmark, it achieves a score of $1.19$, statistically indistinguishable from the Safety expert's $1.18$, whereas Task Arithmetic and LED drift to $2.31$ and $2.41$, indicating compromised safety.





\label{sec:reasoning_and_domain}

\paragraph{ReasonAny ensures domain knowledge preservation and reasoning capability.} 
Table \ref{tab:qwen_7b_bio_performance} illustrates the limitations of standard baselines in domain contexts. 
Methods such as FuseLLM and TIES exhibit catastrophic collapse, indicated by GSM8K scores that drop to negligible levels. 
In contrast, ReasonAny effectively balances capabilities. 
It retains substantial domain expertise with a MedQA score of $47.96$ while preserving logical acuity, evidenced by a GSM8K score of $73.77$ that significantly outperforms the Task Arithmetic baseline. 
Notably, ReasonAny's MMLU score of $73.46$ exceeds both the biomedicine and reasoning models, suggesting the method leverages reasoning logic to enhance domain knowledge application.

More experiments across biomedicine and finance domains can be found in Appendix \ref{app:reasoning_domains_addtional_experiments_bio} and Appendix \ref{app:reasoning_domains_addtional_experiments_fin}, respectively.

\paragraph{ReasonAny performs stably across model families and scales.} 
Moreover, ReasonAny perform stably across Llama3.1 family, results shown in Appendix \ref{app:reasoning_safety_addtional_experiments_llama}.
This stability holds for 14B and 32B models, shown by additional Qwen2.5 experiments in Appendices \ref{app:reasoning_safety_addtional_experiments_qwen_14b},\ref{app:reasoning_safety_addtional_experiments_qwen_32b} and \ref{app:reasoning_safety_addtional_experiments_qwq}.

\begin{table}[t!]
\centering
\caption{Model output word perplexity (PPL) comparison across different merging families: Qwen2.5-7B (Safety, BioMedicine) and Qwen2.5-14B (Safety). The best performance of PPL is highlighted in \textbf{bold}.}
\vspace{-0.2em}
\label{tab:ppl_main}
\resizebox{0.95\linewidth}{!}{%
\begin{tabular}{l | c c c}
\toprule
\textbf{Path} & \textbf{Qwen 7B Safety} & \textbf{Qwen 14B Safety} & \textbf{Qwen 7B Bio.} \\
\midrule
\textbf{Domain Expert}      & 9.32 & 6.63 & 9.14 \\
\textbf{Reasoning}          & 31.25 & 10.63 & 31.25 \\
\hline
\textbf{linear}             & 45.56 & 6.41 & 43.32 \\
\textbf{Task Arithmetic}    & 25.96 & 6.05 & 25.53 \\
\textbf{TIES}               & 2419.98 & 6.75 & 3043.20 \\
\textbf{DARE}               & 505969.92 & 6.31 & 750247.14 \\
\textbf{FuseLLM}            & 44.31 & 6.12 & 34.31 \\
\textbf{LED}                & \textbf{8.79} & \textbf{5.95} & \textbf{8.73} \\
\rowcolor{lightgray!25}
\textbf{ReasonAny}          & 9.32 & 6.08 & 8.82 \\
\bottomrule
\end{tabular}%
}
\vspace{-0.2em}
\end{table}
\label{sec:output_stability}

\paragraph{ReasonAny does not suffer from output collapse.}
\label{sec:output_stability}

ReasonAny ensures functional integrity, effectively avoiding the output collapse observed in baselines. As shown in \textit{italic} with $^*$ mark in Table \ref{tab:qwen_7b_safety_performance}, methods like TIES, DARE, and FuseLLM display a deceptive ``safety'' advantage on HarmBench with 0.02 or 0.00 versus ReasonAny's 0.08. 
However, this anomaly is a artifact of the ``destructive performance collapse'', where these models suffer from collapse in domain-specific performance and lose basic reasoning capabilities, evidenced by their collapse of performance on reasoning benchmarks. 
Since HarmBench relies on a fine-tuned Llama-2-13B classifier to detect harmful content \citep{mazeika2024harmbench}, the incoherent or null outputs produced by these collapsed models fail to trigger the classifier, resulting in artificially low Attack Success Rates (ASR). 
In contrast, ReasonAny maintains reasoning stability as further validated by the low Perplexity (PPL) metrics in Table \ref{tab:ppl_main}, demonstrating its safety scores reflect genuine alignment rather than model failure. 
For more detailed analysis, we provide expanded evaluations across different model scales and domains in Appendix \ref{app:output_content_analysis}.

\subsection{Ablation Study}

\label{sec:ablation_study}


We investigate the contribution of ReasonAny's two key components: \textbf{Reasoning Parameter Identification} and \textbf{Safety Parameter Identification}. 
We conduct ablation studies by selectively removing each module to evaluate their impact on safety and reasoning capabilities, shown in Table \ref{tab:qwen_7b_ablation}. 


Removing {Reasoning Parameter Identification} (w/o reason select) causes a catastrophic collapse in reasoning, with the \textit{GSM8K} score decreasing to $0.15$, confirming that reasoning capabilities rely on preserving specific low-gradient regions. 
Conversely, excluding {Safety Parameter Identification} (w/o safety select) compromises safety, increasing \textit{Safety-Tuned} harmfulness reward to $2.39$ due to the loss of task-specific safety alignment. 
By synthesizing these strategies, ReasonAny maintains a high \textit{GSM8K} score of $86.28$ while minimizing harmfulness reward to $0.84$, suggesting that distinct handling of reasoning and task parameters is essential for building models that are both with reasoning capabilities and safety alignment.


\subsection{Hyperparameter Analysis}
\label{hyper_parameter_analysis}

In this section, we provide ReasonAny's hyperparameter analysis. We evaluate ReasonAny performance with Qwen2.5-7B family models using different scaling factor $\lambda$ and selection ratio $p$.

\begin{figure}[t]
    \centering
    \includegraphics[width=1.0\linewidth]{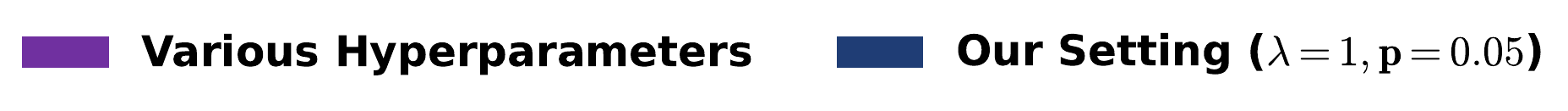}
    \vspace{-1.0em} 
    
    \begin{subfigure}{0.48\linewidth}
        \centering
        \includegraphics[width=\linewidth]{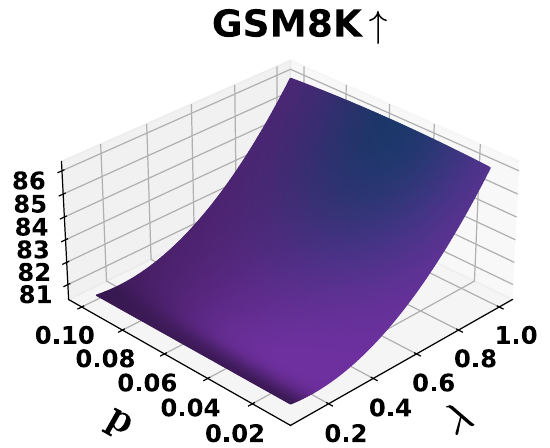}
        \caption{} 
        \label{fig:qwen_hyper_gsm8k}
    \end{subfigure}
    \hfill 
    \begin{subfigure}{0.48\linewidth}
        \centering
        \includegraphics[width=\linewidth]{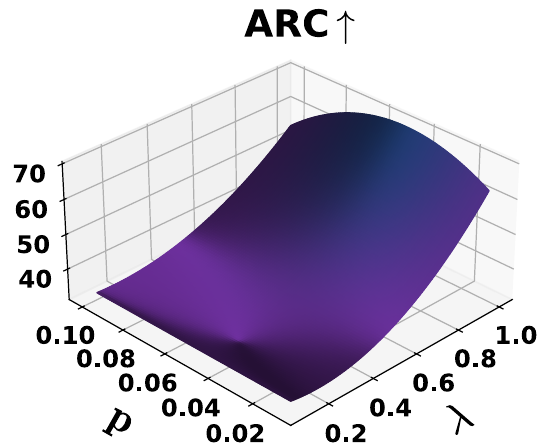}
        \caption{}
        \label{fig:qwen_hyper_arc}
    \end{subfigure}


    \begin{subfigure}{0.48\linewidth}
        \centering
        \includegraphics[width=\linewidth]{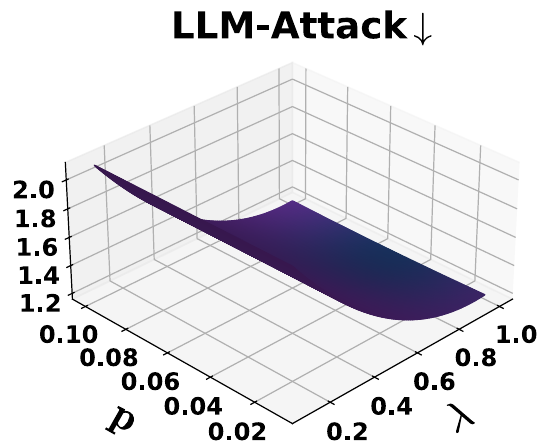}
        \caption{}
        \label{fig:qwen_hyper_llmattack}
    \end{subfigure}
    \hfill
    \begin{subfigure}{0.48\linewidth}
        \centering
        \includegraphics[width=\linewidth]{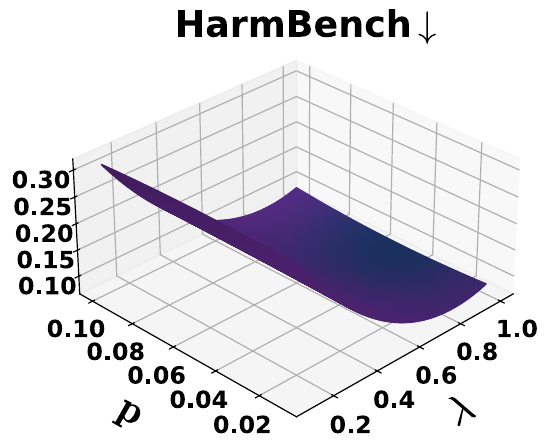}
        \caption{}
        \label{fig:qwen_hyper_harmbench}
    \end{subfigure}
    
    \vspace{0.5em}
    \caption{Hyperparameter analysis of ReasonAny performance across GSM8K (top-left), ARC (top-right), LLM-Attack (bottom-left) and HarmBench (bottom-right) with various scaling factor $\lambda$ and select ratio $p$.}
    \label{fig:hyper_analysis}
\end{figure}

Shown in Figure \ref{fig:hyper_analysis}, we examine the parameter space defined by $\lambda \in \{0.1, 0.5, 1.0\}$ and $p \in \{0.01, 0.05, 0.1\}$, ultimately adopting $\lambda=1.0$ and $p=0.05$ as the optimal configuration. 
3D surface plots smoothly illustrate Qwen2.5-7B’s reasoning-safety performance relative to $\lambda$ and $p$. 
While performance is insensitive to selection ratio $p$, increasing scaling factor $\lambda$ consistently yields monotonic improvements across all datasets.

\vspace{-0.2em}

\section{Related Work}

\vspace{-0.2em}


\subsection{Model merging} 

Model merging is designed to synthesize multiple specialized models into a unified, robust model \citep{goddard-etal-2024-arcees-mergekit, yang2024modelmerge,ruan2023fromtask,li2023deepmodelfusion,lu2024mergeensemblecooperate}, effectively bypassing the need for costly retraining \citep{ilharco2023editing,alexandrov-etal-2024-mitigating}. 
Recent advances mitigate parameter interference and enhance efficiency through methods like TIES \citep{yadav2023tiesmerging}, DARE \citep{yu2024dare}, and related techniques \citep{jin2023dataless, matena2022fisher, wan24fusellm, yu2024dare, liu2025sens}.
The application of model merging has extended to specific areas including cross-lingual transfer \citep{yang2023adamerge}, safety alignment \citep{djuhera2025safemerge, ma2025led, yang2025mitigatingbackdoor}, and pre-training optimization \citep{li2025modelpretraining}. More importantly, merging reasoning models has recently garnered significant attention \citep{zbeeb2025reasoning, pipatanakul2025adapting, hu2025beyond}.
Recent works \citet{tang-etal-2025-unlocking}, \citet{lan2025thinking}, and \citet{yang2025rcp} emphasize merging reasoning models to balance efficiency and depth, notably \citet{yang2025rcp} which utilizes Fisher matrix constraints to prevent reasoning collapse.


\vspace{-0.2em}

\subsection{Neuron-based LLM Interpretation}

\vspace{-0.2em}

Unraveling the internal mechanisms of LLMs is critical for ensuring reliability and building more robust systems \citep{dang2024explainable, wu2024usablexai}. Recent studies have mapped specific capabilities to distinct components, such as domain-specific knowledge, safety and skill neurons \citep{wang-etal-2022-finding-skill, dai-etal-2022-knowledge, christ-etal-2025-mathneurons, zhao2025unravelingjailbreak, qian-etal-2025-tug}. In multilingual settings, proficiency relies on specific neurons in top and bottom layers, while concept representations remain language-agnostic \citep{tang-etal-2024-languageSpecificNeurons, dumas-etal-2025-separating}. Similarly, safety-critical neurons can be calibrated to effectively steer model behaviors like refusal or conformity \citep{zhao2025unravelingjailbreak, wu2024usablexai}. When explainable mechanism meets LLMs' reasoning capability, methods like causal mediation and neuron activation have been used to trace arithmetic processing and explain Chain-of-Thought efficacy \citep{stolfo-etal-2023-mechanistic, rai-yao-2024-investigation, tang-etal-2025-enhancing}. Structural innovations utilize weight and attention interpretation to further optimize these multi-hop processes \citep{punjwani2025weight, yu-etal-2025-back}. Moreover, representation engineering has successfully unlocked reasoning capabilities by isolating specific patterns and parameters \citep{tang-etal-2025-unlocking, christ-etal-2025-mathneurons}. Inspired by gradient-based perspectives on thinking speeds \citep{li-etal-2025-happened}, we deepen the understanding of reasoning evolution through gradient perspective.

\vspace{-0.1em}
\section{Conclusion}
\label{sec:conclusion}

\vspace{-0.1em}

In this paper, we proposed ReasonAny, a model merging framework that aims to merge reasoning models with domain-specific task models. 
We use contrastive gradient identification to take advantage of a key difference: reasoning capabilities are found in parts of the model with small-magnitude gradients, while domain-specific knowledge is found in model weights with large-magnitude gradients.
Experiments demonstrate that ReasonAny significantly outperforms state-of-the-art baselines, preserving both {reasoning} capability and domain-specific task expertise.
\section*{Limitations}
Despite its efficacy, {ReasonAny} has several limitations. First, while the {exclusion} process resolves parameter conflicts, it assumes that reasoning and domain knowledge reside in strictly disjoint subspaces; however, significant overlap in certain complex tasks may still lead to minor interference. 
Second, the current methodology focuses on merging two models (``Reasoning + X''), and its scalability to {multi-model merging} involving several distinct domains remains unexplored. 
Finally, the reliance on {gradient-based attribution} increases the computational overhead during the identification phase compared to simple weight-averaging methods.

\section*{Broader Impact and Ethics Statement}
Our proposed framework, ReasonAny, significantly advances the efficiency of Large Language Model development by enabling the training-free synthesis of reasoning and domain-specific capabilities, thereby reducing the computational resources and carbon footprint associated with retraining. 
Crucially, our experiments demonstrate that ReasonAny effectively preserves safety alignment parameters, mitigating the risks of jailbreaking or safety degradation often observed in other model merging techniques. 
However, the deployment of enhanced reasoning models in high-stakes domains, such as biomedicine and finance, necessitates caution. 
We strongly advise that such models be used with rigorous human oversight to address potential biases inherited from source models and to prevent over-reliance on automated decision-making in critical scenarios.


\bibliography{custom}

\clearpage
\appendix

\label{sec:appendix}
\section{Theoretical Justification for Gradient Magnitude Metrics}
\label{app:gradients_nuclear_norm_mad}
In this section, we provide the mathematical derivation explaining the relationship between the Nuclear Norm, Mean Absolute Difference (MAD) across layers, and the intrinsic magnitude of the model gradients. This derivation theoretically grounds the methodology used in Section \ref{sec:low_gradients}, specifically justifying why lower values of nuclear norm and MAD are positively correlated with smaller gradient updates (characteristic of reasoning capabilities).

\subsection{Relationship between Nuclear Norm and Gradient Magnitude}

\label{app:gradients_nuclear_norm}

Let $G_{X,l} \in \mathbb{R}^{m \times n}$ denote the gradient matrix for a specific projection layer $X \in \{Q, K, V, O\}$ at layer index $l$. The magnitude of the parameter update is typically quantified by the Frobenius norm $\|G_{X,l}\|_F$, which corresponds to the Euclidean norm of the flattened gradient vector:
\begin{equation}
    \|G_{X,l}\|_F = \sqrt{\sum_{i=1}^{\min(m,n)} \sigma_i^2}
\end{equation}
where $\sigma_i$ are the singular values of $G_{X,l}$.

The nuclear norm $s_{X,l}$ is utilized as our primary metric, is defined as the sum of the singular values (the $\ell_1$ norm of the spectrum):
\begin{equation}
    s_{X,l} = \|G_{X,l}\|_* = \sum_{i=1}^{\min(m,n)} \sigma_i
\end{equation}

To establish the positive correlation between the nuclear norm and the gradient magnitude, we invoke the standard norm inequalities. For any matrix $A$ of rank $r$, the relationship between the Frobenius norm and the nuclear norm is given by:
\begin{equation}
    \|G_{X,l}\|_F \leq \|G_{X,l}\|_* \leq \sqrt{r} \|G_{X,l}\|_F
\end{equation}
The left inequality $\|G_{X,l}\|_F \leq s_{X,l}$ is crucial. It implies that the nuclear norm serves as a strictly convex upper bound on the Frobenius norm. Therefore, minimizing the nuclear norm ($s_{X,l} \to 0$) mathematically necessitates the minimization of the Frobenius norm ($\|G_{X,l}\|_F \to 0$).

Consequently, a smaller nuclear norm directly implies a smaller gradient magnitude. This justifies the observation that the ``reasoning'' subspace, characterized by \textbf{low nuclear norms}, resides in the \textbf{low-gradient model weights}.

\begin{figure}[t!]
    \centering
    
    

    \begin{subfigure}{0.48\linewidth}
        \centering
        \includegraphics[width=\linewidth]{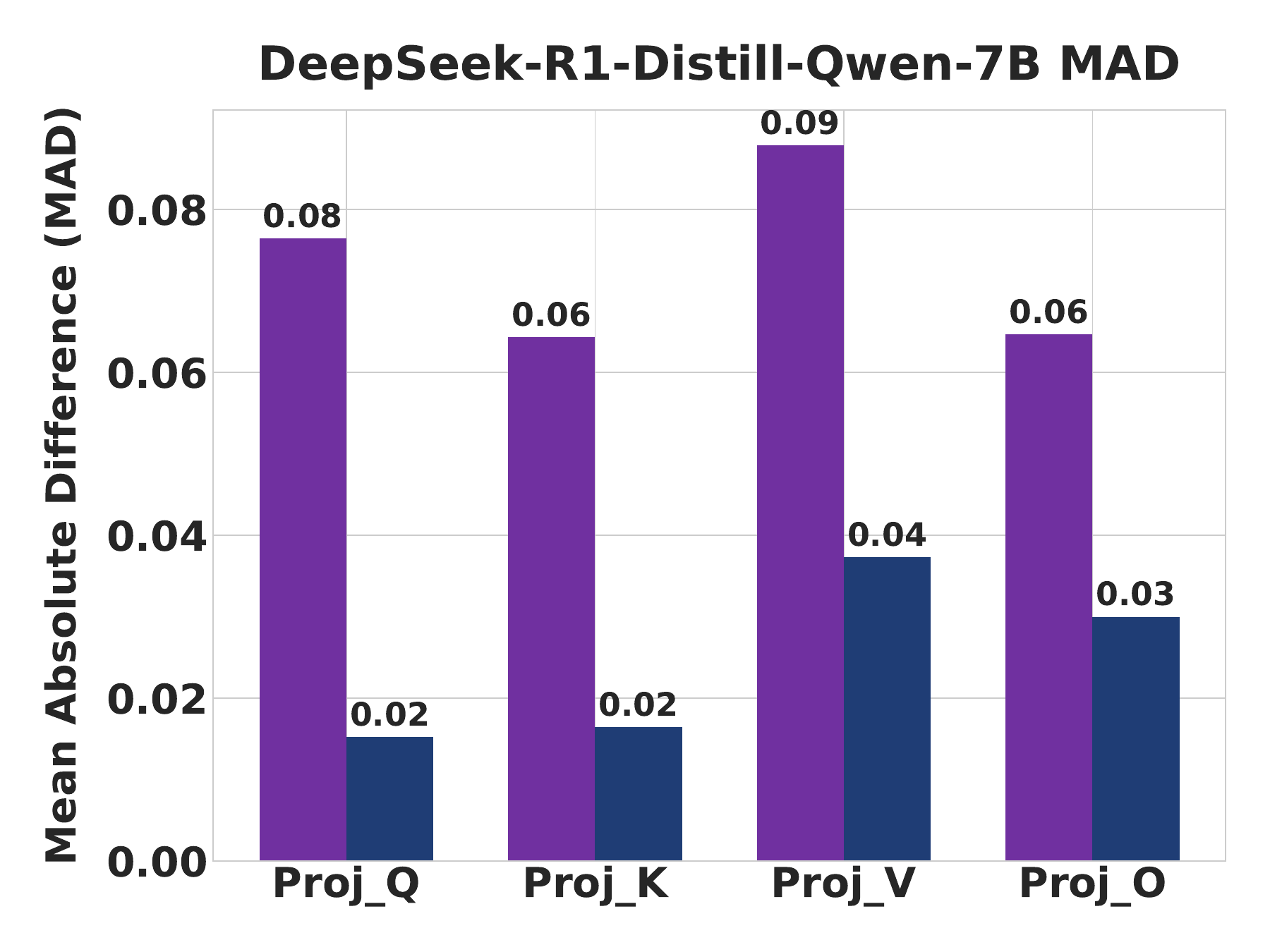}
        \caption{} 
        \label{fig:mad_r1_qwen_7b}
    \end{subfigure}
    \hfill
    \begin{subfigure}{0.48\linewidth}
        \centering
        \includegraphics[width=\linewidth]{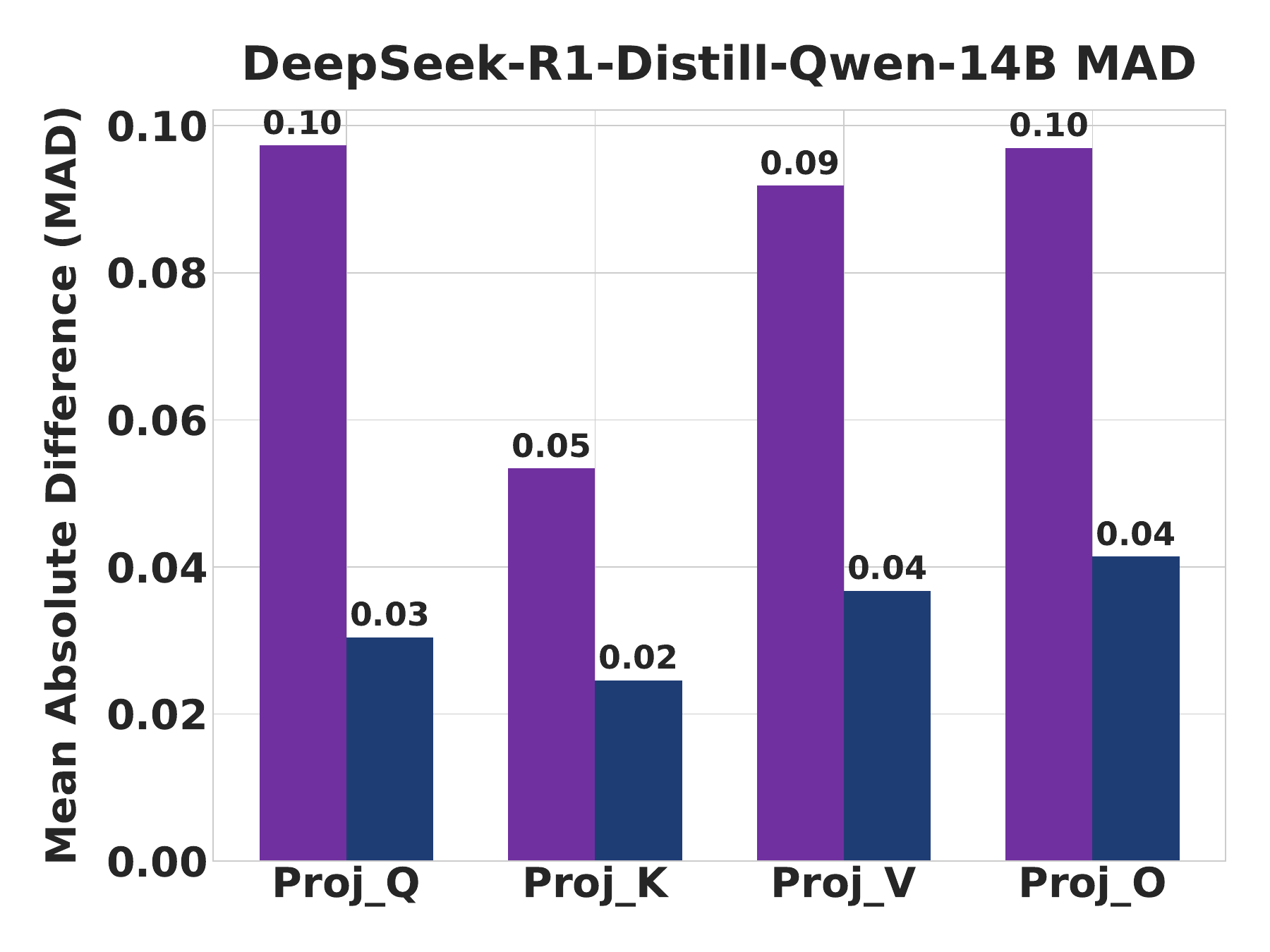}
        \caption{}
        \label{fig:mad_r1_qwen_14b}
    \end{subfigure}
    
    \vspace{0.2em}
    \caption{The left and left panel illustrates the Mean Absolute Difference (MAD) for Qwen2.5-7B and Qwen2.5-14B, quantifying the average magnitude difference across layers.}
    \label{fig:mad}
    \vspace{-0.7em}
\end{figure}
\subsection{Gradient Stability Analysis}
\label{subsec:amd_analysis}
To assess the layer-wise stability of these updates, we employ the \textbf{Mean Absolute Difference} (MAD):
\begin{equation}
    \text{MAD} s_x = \frac{1}{N-1} \sum_{i=1}^{N-1} |s_{x,i+1} - s_{x,i}|.
\end{equation}

As shown in Figure \ref{fig:mad} for Qwen2.5-7B and Qwen2.5-14B series models, DeepSeek-R1-Distill reasoning models, as the blue line marked as long-CoT in the figure, exhibit significantly \textbf{lower nuclear norms} than the purple line marked as Vanilla-CoT. 
Furthermore, as quantified by the MAD scores in the bottom-right subfigure, these reasoning models demonstrate \textbf{higher stability} across layers compared to the high-magnitude fluctuations observed in standard task fine-tuning.
\subsection{Connection between Gradient Magnitude and MAD}

\label{app:gradients_mad}


We demonstrate that a globally small gradient magnitude implies a small MAD. Assume that the gradient magnitude is bounded by a small constant $\epsilon$ across all layers, such that $0 \le s_{X,l} \le \epsilon$ for all $l$. By the triangle inequality, the difference between any two layers is bounded by:
\begin{equation}
    |s_{X, l+1} - s_{X, l}| \leq \max(s_X) - \min(s_X) \leq \epsilon
\end{equation}
Substituting this into the definition of MAD:
\begin{equation}
    \text{MAD}(s_X) \leq \frac{1}{N-1} \sum_{l=1}^{N-1} \epsilon = \epsilon
\end{equation}

Thus, as the model gradient becomes smaller, shown as decreasing $\epsilon$, the MAD score is mathematically constrained to decrease. 
This confirms that the ``low-gradient'' model weights identified in reasoning tasks will naturally exhibit both low nuclear norms as small magnitude and \textbf{low MAD values as high stability}. This distinguish them from the high-magnitude, high-fluctuation updates observed in standard knowledge injection.
\section{Baselines Explanation}
\label{app:baselines_explanation}

We detail the model merging baselines employed in our experiments below. We summarize the core formulation and theoretical motivation for each method.

\begin{itemize}
    \item \textbf{Linear} \citep{Izmailov2018Averaging}: This foundational approach performs element-wise averaging of model parameters. It assumes linear interpolation to generalize across tasks.
    \item \textbf{Task Arithmetic} \citep{ilharco2023editing}: This method steers model behavior by manipulating task vectors, defined as the element-wise difference between fine-tuned and pre-trained weights. These vectors are linearly scaled and aggregated to combine distinct task capabilities.
    \item \textbf{TIES-Merging} \citep{yadav2023tiesmerging}: Designed to mitigate parameter interference, TIES reduces redundancy by retaining only the top-$k$ magnitude updates (Trim). It subsequently resolves sign conflicts among models (Elect) before aggregating the unified signs (Merge).
    \item \textbf{DARE-Merging} \citep{yu2024dare}: DARE approximates the original model's topology by stochastically pruning delta parameters (Drop) and rescaling the remaining weights (Rescale). This reduces the magnitude of parameter shifts while preserving task-specific functional improvements.
    \item \textbf{FuseLLM} \citep{wan24fusellm}: Distinct from direct weight manipulation, FuseLLM leverages knowledge fusion by aligning the merged model's token probability distributions with those of the source LLMs, minimizing the Kullback-Leibler divergence to preserve capabilities.
    \item \textbf{LED-Merging} \citep{ma2025led}: LED-Merging addresses safety-utility conflicts by targeting neuron misidentification and cross-task interference. It operates in three stages: {Location} identifies critical neurons via gradient-based attribution; {Election} dynamically selects neurons significant to both base and fine-tuned models; and {Disjoint} isolates conflicting updates through set difference operations to prevent destructive parameter collisions.
\end{itemize}
\begin{table*}[t!]
    \centering
    \renewcommand{\arraystretch}{1.5}
    \small
    \caption{Overview of all evaluation 15 datasets categorized into five primary sub-areas: Reasoning, Knowledge, Safety, Biomedicine, and Finance.}
    \label{tab:all_datasets}
    \resizebox{\textwidth}{!}{
         \begin{tabular}{p{2cm} |p{2cm}|p{2cm}| p{2cm}| p{3cm}| p{4.5cm}}
            \hline
            \textbf{Dataset} &\textbf{Sub Area Type}& \textbf{Question Type} & \textbf{Metric} & \textbf{Category} & \textbf{Explanation} \\
            \hline
            \textbf{GSM8K} \citep{Cobbe21gsm} &Reasoning & Numerical Math & Accuracy & Math \& Reasoning & High-quality grade school math word problems requiring multi-step reasoning with basic arithmetic. \\ \hline
            \textbf{MATH} \citep{lightman2023letsmath500}&Reasoning & Numerical Math & Numerical Accuracy $\uparrow$ & Math \& Reasoning & Comprehensive dataset of 500 challenging competition-level math problems across seven subject areas. \\ \hline
            \textbf{AIME24} \citep{aime_1983_2024}& Reasoning& Numerical Math & Numerical Accuracy $\uparrow$ & Math \& Reasoning & Problems from the 2024 AIME, evaluating reasoning capabilities on fresh, uncontaminated data. \\
            \hline 
            \textbf{HumanEval} \citep{Mark21humaneval}&Reasoning & Code Generation & Pass@1 $\uparrow$ & Code \& Reasoning & 164 hand-written Python problems evaluating functional correctness through function signatures and unit tests. \\ \hline
            \textbf{LiveCodeBench} \citep{jain2024livecodebench} &Reasoning& Code Generation & Pass@1 $\uparrow$ & Code \& Reasoning & Contest problems released after training cutoff to assess generalization and prevent data contamination. \\
            \hline
\textbf{ARC} \citep{Clark18arc} &Knowledge& 
Single Choice Question& 
Single Choice Question Accuracy $\uparrow$ & 
Knowledge QA & 
Grade-school science questions (Easy/Challenge) requiring complex reasoning and knowledge integration; designed to resist simple retrieval and co-occurrence statistics. \\
\hline
\textbf{MMLU} \citep{hendryckstest2021mmlu1,hendrycks2021ethicsmmlu2}&Knowledge& 
Single Choice Question& 
Single Choice Question Accuracy $\uparrow$ & 
Knowledge QA \& Scientific Reasoning & 
Comprehensive benchmark measuring multitask accuracy across 57 subjects (STEM, humanities, etc.) to assess general world knowledge and problem-solving capabilities. \\
\hline
\textbf{GPQA} \citep{rein2023gpqa}&Knowledge& 
Single Choice Question& 
Single Choice Question Accuracy $\uparrow$ & 
Knowledge QA \& Scientific Reasoning & 
Challenging graduate-level biology, physics, and chemistry questions written by experts. ``Google-proof'' design tests scientific reasoning difficult to solve via search. \\
\hline
\textbf{LLM-Attack} \citep{zou2023universalllm-attack} &Safety & Malicious Question & Deberta-V3 Redteam Model Evaluation Score $\downarrow$ & Safety & Uses AdvBench to test adversarial suffixes optimized via Greedy Coordinate Gradient for affirmative harmful responses, lower scores indicating better safety alignment. \\ \hline
\textbf{HarmBench} \citep{mazeika2024harmbench} &Safety & Malicious Question & HarmBench-Llama-2-13b Attack Success Rate (ASR $\downarrow$) & Safety & Standardized framework with 510 behaviors across multiple categories using a fine-tuned classifier for attack rate assessment, lower scores indicating better safety alignment. \\ \hline
\textbf{SafeChain} \citep{jiang2025safechain}&Safety & Vanilla \& Malicious Question & OpenAI o4-mini Model Evaluation Score $\uparrow$& Safety \& Reasoning & Evaluates safety within Chain-of-Thought traces while preserving reasoning utility using o4-mini ranged from 0.00 to 5.00, the higher score indicates safer reasoning process. \\
\hline
\textbf{PubMedQA} \citep{Jin19PubMedQA} &Biomedicine & Single Choice Question& Single Choice Question Accuracy $\uparrow$ & Knowledge QA \& Biomedicine & Biomedical dataset answering research questions (yes/no/maybe) using abstracts, requiring reasoning over quantitative findings in the text. \\ \hline
\textbf{MedQA} \citep{Jin20MedQA} &Biomedicine & Single Choice Question& Single Choice Question Accuracy $\uparrow$ & Knowledge QA \& Biomedicine & Open-domain multiple-choice dataset from US, China, and Taiwan medical exams, testing professional clinical knowledge and complex reasoning. \\ 
\hline
\textbf{ConvFinQA} \citep{chen2022convfinqa} &Finance & Numerical Finance Problem & Numerical Accuracy $\uparrow$ & Finance Calculation & Focuses on numerical reasoning chains in conversational QA over financial reports, requiring complex calculations on text and tables. \\ \hline
\textbf{OpenFinData} \citep{openfindata} &Finance & Single Choice Question & Single Choice Question Accuracy $\uparrow$ & Knowledge QA \& Finance & Comprehensive benchmark with six modules covering calculation, analysis, and compliance, derived from authentic industrial financial scenarios. \\
\hline
        \end{tabular}
    }
\end{table*}
\section{Datasets Explanation}
\label{app:datasets_explanation}
\subsection{Evaluation Datasets}
To comprehensively evaluate the capabilities of our merged models, we employ a diverse benchmark suite comprising 15 datasets categorized into five primary sub-areas: Reasoning, Knowledge, Safety, Biomedicine, and Finance. Detailed specifications for each dataset are provided in Table \ref{tab:all_datasets}.

We assess mathematical and algorithmic \textbf{Reasoning} capabilities using GSM8K, MATH, AIME24, HumanEval, and LiveCodeBench. For general world \textbf{Knowledge} and scientific understanding, we utilize ARC, MMLU, and the graduate-level GPQA benchmark. To ensure robust alignment, we evaluate \textbf{Safety} using LLM-Attack, HarmBench, and the reasoning-focused SafeChain. Finally, we examine domain generalization through specialized datasets in \textbf{Biomedicine} (PubMedQA, MedQA) and \textbf{Finance} (ConvFinQA, OpenFinData). This multi-faceted evaluation strategy allows us to verify that improvements in reasoning do not come at the cost of safety or general knowledge retention.

\label{app:datasets_explanation_performance_bench}

\label{app:datasets_explanation_reasoning_bench}

\label{app:datasets_explanation_knowledge_bench}

\label{app:datasets_explanation_safety_bench}

\label{app:datasets_explanation_domain_bench}

\label{app:datasets_explanation_bio_bench}

\label{app:datasets_explanation_fin_bench}

\subsection{Calibration Datasets}
To precisely isolate task-specific subspaces using Contrastive Gradient Identification, we employ diverse calibration datasets representing distinct capabilities. 
Specifically, we utilize: 
(1) \textit{OpenThoughts-114k-math} \citep{openr1_2025_openthoughts}\footnote{\url{https://huggingface.co/datasets/open-r1/OpenThoughts-114k-math}} for the \textbf{reasoning} domain; 
(2) \textit{hh-rlhf} \citep{Bai2022TrainingAHhh-rlhf}\footnote{\url{https://huggingface.co/datasets/Anthropic/hh-rlhf}} for \textbf{safety} constraints; 
(3) \textit{PubMedQA} \citep{Jin19PubMedQA}\footnote{\url{https://huggingface.co/datasets/qiaojin/PubMedQA}} for \textbf{biomedicine}; 
and (4) \textit{FinanceQA} \citep{Mateega2025FinanceQAAB}\footnote{\url{https://huggingface.co/datasets/AfterQuery/FinanceQA}} for \textbf{finance}.

To ensure the high reproducibility of {ReasonAny} and minimize computational overhead during the gradient attribution phase, we select only the first 100 samples from each dataset to form the calibration sets. These samples serve as the representative distribution to compute the contrastive scores, allowing the framework to efficiently identify the topologically distinct parameter regions associated with either long-CoT reasoning or domain expertise.
\section{Full Models Configuration}
\label{app:models_config}

\subsection{Safety Evaluation}
We conduct comprehensive experiments across the Qwen2.5 \citep{qwen2.5} and Llama-3.1 \citep{grattafiori2024llama} model families to evaluate the synthesis of safety and reasoning. For the base models, we utilize Qwen2.5-\{7B, 14B, 32B\}\footnote{\url{https://huggingface.co/Qwen/Qwen2.5-7B}} and Llama-3.1-8B\footnote{\url{https://huggingface.co/meta-llama/Llama-3.1-8B}}.
To construct specialized Safety experts, we apply Low-Rank Adaptation (LoRA) fine-tuning \citep{hu2022lora,wang2023alpacalora} to the instruction-tuned variants of these backbones on Safety-Tuned LLaMAs dataset \citep{bianchi2024safetytuned} to obtain the best safety alignment performance models on certain base model.
For the Reasoning experts, we employ state-of-the-art distilled reasoning models, specifically DeepSeek-R1-Distill-Qwen-\{7B, 14B, 32B\}\footnote{\url{https://huggingface.co/deepseek-ai/DeepSeek-R1-Distill-Qwen-7B}} and DeepSeek-R1-Distill-Llama-8B\footnote{\url{https://huggingface.co/deepseek-ai/DeepSeek-R1-Distill-Llama-8B}} \citep{guo2025deepseek}. Additionally, at the 32B scale, we incorporate QwQ-32B-Preview\footnote{\url{https://huggingface.co/Qwen/QwQ-32B-Preview}} \citep{qwq32b} to verify the framework's generalization across different reasoning architectures.

\subsection{Domain Evaluation}

\subsubsection{Biomedicine Evaluation}
In the biomedical domain, we utilize Qwen2.5-7B and Llama-3.1-8B as foundational backbones. The domain-specific experts are Meditron3-Qwen2.5-7B\footnote{\url{https://huggingface.co/OpenMeditron/Meditron3-Qwen2.5-7B}} \citep{Chen23Meditron} and MMed-Llama-3-8B\footnote{\url{https://huggingface.co/Henrychur/MMed-Llama-3-8B}} \citep{qiu2024building-mmed-llama}, selected for their extensive medical pre-training. These are merged with their corresponding DeepSeek-R1 distilled reasoning models, DeepSeek-R1-Distill-Qwen-7B and DeepSeek-R1-Distill-Llama-8B, to assess the preservation of clinical knowledge alongside reasoning capability.

\subsubsection{Finance Evaluation}
For financial reasoning tasks, we adopt the WiroAI series models as the domain-specific experts. Specifically, we employ WiroAI-Finance-Qwen-7B\footnote{\url{https://huggingface.co/WiroAI/WiroAI-Finance-Qwen-7B}} \citep{WiroAI-finance-qwen-7b} paired with the Qwen2.5-7B base, and WiroAI-Finance-Llama-8B\footnote{\url{https://huggingface.co/WiroAI/WiroAI-Finance-Llama-8B}} \citep{WiroAI-finance-llama-8b} paired with the Llama-3.1-8B base. These models are merged with DeepSeek-R1-Distill-Qwen-7B and DeepSeek-R1-Distill-Llama-8B, respectively, to evaluate the synergistic integration of financial literacy and logical reasoning capabilities.
\section{Experiment Setting of Figure \ref{fig:intro_fig}}
\label{sec:exp_fig1_setting}

The experimental settings for Figure \ref{fig:intro_fig} are identical to those described in Section \ref{subsec:setup}. 
We utilize the GSM8K dataset \citep{Cobbe21gsm} and the Safety-Tuned dataset \citep{bianchi2024safetytuned} to represent the performance of different model merging methods on reasoning and domain-specific tasks, respectively. Regarding the safety metric, while the Safety-Tuned benchmark \citep{bianchi2024safetytuned} originally yields a harmfulness score, we follow the methodology of Safety-Tuned to convert this into a Safety Score with maximum score of 4.0. 
Specifically, we subtract the current harmfulness score from the maximum possible score as \textit{$\text{Safety Score} = \text{Max Score} - \text{Harmfulness Score}$}, ensuring that higher scores correspond to better safety performance in our visualization.
\section{Merging Methods Hyperparameter Setting}
\label{app:hyper_setting}

Utilizing the mergekit repository\footnote{\url{https://github.com/arcee-ai/mergekit}} \citep{goddard-etal-2024-arcees-mergekit}, for baseline methods, we apply the following hyperparameter. In Task Arithmetic, the scaling factor is set to $\lambda=0.3$. For both TIES-Merging and DARE, the merging weight is $\lambda=0.3$ and the dropout rate is $r=0.9$. For LED-Merging\footnote{\url{https://github.com/MqLeet/LED-Merging}}, we utilize the ratio for selection with 0.1 and the scaling term $\lambda$ of 1.0.
For ReasonAny, the model weight selection ratio $p_r$ is set to 0.05 for both reasoning and task model and the scaling factor is set to 1.0 for optimal performance.

During inference, we set `max\_new\_tokens' to 4096 and `temperature' to 0 for the base and task models. For the reasoning model, we use `max\_new\_tokens' of 32768, `temperature' of 0.6, and `top-k' of 0.95 for long-CoT generation. 

\begin{table*}[t!]
\centering
\caption{Performance comparison of merging Qwen2.5-14B family with safety fine-tuning Qwen2.5-14B-Instruct (Safety) and DeepSeek-R1-Distill-Qwen-14B (Reasoning) on all datasets across Reasoning, Knowledge and Safety Benchmarks, where \textbf{Average} $\uparrow$ column indicate average performance across performance bench. The best performance among all merging methods on each dataset is highlighted in \textbf{bold}.}
\label{tab:qwen_14b_safety_performance}
\renewcommand{\arraystretch}{1.1} 
\resizebox{\textwidth}{!}{
\begin{tabular}{l|ccccccccc|c|ccc}

\hline

{\textbf{Eval Bench}} & \multicolumn{10}{c|}{\textbf{Performance Bench}} & \multicolumn{3}{c}{\textbf{Safety Bench}} \\

\cline{1-14}

\textbf{Sub Areas}& \multicolumn{5}{c|}{\textbf{Reasoning}} & \multicolumn{4}{c|}{\textbf{Knowledge}}& \multicolumn{1}{c|}{\textbf{}} & \multicolumn{3}{c}{\textbf{Safety}} \\

\cline{1-14}

\textbf{Datasets} & \textbf{GSM8K}$\uparrow$ & \textbf{Math}$\uparrow$ & \textbf{AIME}$\uparrow$ & \textbf{HumanEval}$\uparrow$ & \textbf{LiveCodeBench}$\uparrow$ & \textbf{ARC-C}$\uparrow$ & \textbf{ARC-E}$\uparrow$ & \textbf{MMLU}$\uparrow$ & \textbf{GPQA}$\uparrow$ & \textbf{Average}$\uparrow$ & \textbf{Safety-Tuned}$\downarrow$ & \textbf{HarmBench}$\downarrow$ & \textbf{SafeChain}$\uparrow$ \\

\hline
\textbf{Safety} & 75.74 & 76.60 & 20.00 & 78.80 & 27.31 & \textbf{92.21} & 97.18 & \textbf{78.73} & 39.39 & 65.11 & \textbf{1.10} & \textbf{0.06} & \textbf{4.84} \\
\textbf{Reasoning} & \textbf{86.43} & \textbf{92.40} & \textbf{63.33} & \textbf{95.73} & \textbf{48.15} & 88.41 & \textbf{99.50} & 73.28 & \textbf{57.10} & \textbf{78.37} & 1.66 & 0.21 & 4.56 \\
\hline
\textbf{Linear} & 50.57 & 80.80 & 13.33 & 89.02 & 13.33 & 91.97 & 97.35 & 77.63 & 35.61 & 61.07 & 1.27 & 0.10 & 4.47 \\
\textbf{Task Arithmetic} & 81.96 & 81.00 & 36.67 & 61.59 & 39.27 & 88.47 & 97.53 & 78.79 & 47.73 & 68.11 & 1.42 & 0.09 & 4.25 \\
\textbf{TIES} & 79.76 & 70.00 & 6.67 & 84.15 & 30.81 & 85.87 & 96.83 & 72.93 & 36.36 & 62.60 & {2.46} & 0.56 & 4.25 \\
\textbf{DARE} & 64.90 & 77.80 & 16.67 & 64.63 & 8.34 & 91.59 & 91.53 & 76.94 & 41.23 & 59.29 & 1.29 & {0.03} & \textbf{4.80} \\
\textbf{FuseLLM} & 52.46 & 75.40 & 10.00 & 19.51 & 14.48 & 91.29 & 97.71 & 76.25 & 28.78 & 51.76 & {2.53} & 0.29 & 4.52 \\
\textbf{LED} & 76.42 & 76.60 & 30.00 & 40.85 & 30.35 & 92.22 & 97.18 & 77.56 & {53.75} & 63.88 & 1.84 & 0.35 & 4.47 \\
\rowcolor{lightgray!25} \textbf{ReasonAny} & \textbf{85.44} & \textbf{83.20} & \textbf{46.67} & \textbf{92.32} & \textbf{44.35} & \textbf{92.52} & \textbf{97.71} & \textbf{78.86} & \textbf{57.50} & \textbf{75.40} & \textbf{1.10} & \textbf{0.03} & 4.77 \\
\hline
\end{tabular}}
\end{table*}
\begin{table*}[t!]
\centering
\caption{Performance comparison of merging Qwen2.5-32B family with safety fine-tuning Qwen2.5-32B-Instruct (Safety) and DeepSeek-R1-Distill-Qwen-32B (Reasoning) on all datasets across Reasoning, Knowledge and Safety Benchmarks. The best performance among all merging methods on each dataset is highlighted in \textbf{bold}.}
\label{tab:qwen_32b_safety_performance}
\renewcommand{\arraystretch}{1.2}
\resizebox{\textwidth}{!}{
\begin{tabular}{l|ccccccccc|c|ccc}

\hline

{\textbf{Eval Bench}} & \multicolumn{10}{c|}{\textbf{Performance Bench}} & \multicolumn{3}{c}{\textbf{Safety Bench}} \\

\cline{1-14}

\textbf{Sub Areas}& \multicolumn{5}{c|}{\textbf{Reasoning}} & \multicolumn{4}{c|}{\textbf{Knowledge}}& \multicolumn{1}{c|}{\textbf{}} & \multicolumn{3}{c}{\textbf{Safety}} \\

\cline{1-14}

\textbf{Datasets} & \textbf{GSM8K}$\uparrow$ & \textbf{Math}$\uparrow$ & \textbf{AIME}$\uparrow$ & \textbf{HumanEval}$\uparrow$ & \textbf{LiveCodeBench}$\uparrow$ & \textbf{ARC-C}$\uparrow$ & \textbf{ARC-E}$\uparrow$ & \textbf{MMLU}$\uparrow$ & \textbf{GPQA}$\uparrow$ & \textbf{Average}$\uparrow$ & \textbf{Safety-Tuned}$\downarrow$ & \textbf{HarmBench}$\downarrow$ & \textbf{SafeChain}$\uparrow$ \\

\hline
\textbf{Safety} & 83.02 & 82.2 & 23.33 & 84.31 & 23.39 & \textbf{95.59} & 98.94 & \textbf{81.78} & 41.67 & 68.25 & \textbf{1.2} & \textbf{0.04} & \textbf{4.83} \\
\textbf{Reasoning} & \textbf{94.90} & \textbf{94.2} & \textbf{73.33} & \textbf{92.41} & \textbf{54.25} & 94.7 & \textbf{99.54} & 79.65 & \textbf{59.39} & \textbf{82.49} & 1.53 & 0.26 & 4.65 \\
\hline
\textbf{Linear} & 82.34 & 81.2 & 16.67 & 82.32 & 17.55 & 95.25 & 97.45 & 81.9 & 38.64 & 65.92 & 1.33 & 0.06 & 4.6 \\
\textbf{Task Arithmetic} & 78.39 & 82.8 & 30 & 88.41 & 19.75 & 95.59 & 97.22 & 81.89 & 36.36 & 67.82 & 1.35 & 0.11 & 4.76 \\
\textbf{TIES} & 91.51 & 76.2 & 33.33 & \textbf{91.46} & 21.86 & 95.93 & 98.59 & 79.52 & 34.09 & 69.17 & 2.31 & 0.46 & 4.42 \\
\textbf{DARE} & 87.87 & 83.4 & 23.33 & 82.32 & 23.78 & 96.27 & 98.59 & 81.85 & 38.64 & 68.45 & 1.29 & 0.06 & 4.81 \\
\textbf{FuseLLM} & 88.55 & 79.6 & 23.33 & 74.39 & 25.89 & 95.59 & 98.59 & 80.62 & 33.33 & 66.65 & 1.97 & 0.35 & 4.76 \\
\textbf{LED} & 69.75 & 78.4 & 20 & 61.59 & 38.6 & 95.88 & 96.88 & 80.83 & 46.97 & 65.43 & 1.57 & 0.33 & 4.58 \\
\rowcolor{lightgray!25}  \textbf{ReasonAny} & \textbf{91.53} & \textbf{92.4} & \textbf{56.67} & 86.59 & \textbf{46.16} & \textbf{96.43} & \textbf{98.75} & \textbf{82.09} & \textbf{56.25} & \textbf{78.54} & \textbf{1.23} & \textbf{0.04} & \textbf{4.86} \\
\hline
\end{tabular}}
\end{table*}
\begin{table*}[t!]
\centering
\caption{Performance comparison of merging Qwen2.5-32B family with safety fine-tuning Qwen2.5-32B-Instruct (Safety) and QwQ-32B-Preview (Reasoning) on all datasets across Reasoning, Knowledge and Safety Benchmarks. The best performance among all merging methods on each dataset is highlighted in \textbf{bold}.}
\label{tab:32b_qwq_safety_performance}
\renewcommand{\arraystretch}{1.2}
\resizebox{\textwidth}{!}{
\begin{tabular}{l|ccccccccc|c|ccc}

\hline

{\textbf{Eval Bench}} & \multicolumn{10}{c|}{\textbf{Performance Bench}} & \multicolumn{3}{c}{\textbf{Safety Bench}} \\

\cline{1-14}

\textbf{Sub Areas}& \multicolumn{5}{c|}{\textbf{Reasoning}} & \multicolumn{4}{c|}{\textbf{Knowledge}}& \multicolumn{1}{c|}{\textbf{}} & \multicolumn{3}{c}{\textbf{Safety}} \\

\cline{1-14}

\textbf{Datasets} & \textbf{GSM8K}$\uparrow$ & \textbf{Math}$\uparrow$ & \textbf{AIME}$\uparrow$ & \textbf{HumanEval}$\uparrow$ & \textbf{LiveCodeBench}$\uparrow$ & \textbf{ARC-C}$\uparrow$ & \textbf{ARC-E}$\uparrow$ & \textbf{MMLU}$\uparrow$ & \textbf{GPQA}$\uparrow$ & \textbf{Average}$\uparrow$ & \textbf{Safety-Tuned}$\downarrow$ & \textbf{HarmBench}$\downarrow$ & \textbf{SafeChain}$\uparrow$ \\

\hline
\textbf{Safety} & 83.02 & 82.20 & 23.33 & 84.31 & 23.39 & 93.48 & 98.24 & \textbf{81.78} & 41.67 & 67.94 & \textbf{1.20} & \textbf{4.83} & \textbf{4.83} \\
\textbf{Reasoning} & \textbf{95.41} & \textbf{84.40} & \textbf{53.33} & \textbf{89.63} & \textbf{57.25} & \textbf{95.25} & \textbf{99.32} & 79.84 & \textbf{53.30} & \textbf{78.64} & 1.72 & 4.64 & 4.64 \\
\hline
\textbf{Linear} & 28.28 & 15.20 & 10.00 & 89.02 & 24.74 & 95.93 & 98.94 & 81.85 & 45.18 & 54.35 & 1.28 & 4.59 & 4.59 \\
\textbf{Task Arithmetic} & 89.84 & 73.60 & 20.00 & 83.54 & 25.02 & 95.59 & 98.77 & 81.76 & 36.36 & 67.16 & 1.40 & 4.79 & 4.79 \\
\textbf{TIES} & 64.67 & 71.40 & 20.00 & 88.41 & \textbf{54.23} & 94.34 & 96.88 & 80.65 & 40.91 & 67.94 & 1.32 & \textbf{4.81} & \textbf{4.81} \\
\textbf{DARE} & 86.20 & 81.00 & 30.00 & 81.32 & 27.13 & 96.27 & 98.59 & 81.85 & 29.94 & 68.03 & 1.28 & 4.60 & 4.60 \\
\textbf{FuseLLM} & 81.73 & 79.80 & 26.67 & 71.34 & 39.85 & 95.25 & 98.59 & 80.80 & 29.94 & 67.11 & 1.73 & 4.46 & 4.46 \\
\textbf{LED} & 69.75 & 78.40 & 20.00 & 84.31 & 48.60 & 95.59 & 98.41 & 80.83 & 52.28 & 69.80 & 1.57 & 4.60 & 4.60 \\
\textbf{ReasonAny} & \textbf{91.13} & \textbf{81.00} & \textbf{36.67} & \textbf{89.71} & \textbf{53.39} & \textbf{96.88} & \textbf{98.94} & \textbf{82.00} & \textbf{55.33} & \textbf{76.12} & \textbf{1.20} & \textbf{4.80} & \textbf{4.80} \\
\hline
\end{tabular}}
\end{table*}
\begin{table*}[ht!]
\centering
\caption{Performance comparison of merging Llama-3.1-8B family with safety fine-tuning Llama-3.1-8B-Instruct (Safety) and DeepSeek-R1-Distill-Llama-8B (Reasoning) on all datasets across Reasoning, Knowledge and Safety Benchmarks, where \textbf{Average} $\uparrow$ column indicate average performance across performance bench. The best performance among all merging methods on each dataset is highlighted in \textbf{bold}.}
\label{tab:llama_8b_safety_performance}
\renewcommand{\arraystretch}{1.1}
\resizebox{\textwidth}{!}{
\begin{tabular}{l|ccccccccc|c|ccc}

\hline

{\textbf{Eval Bench}} & \multicolumn{10}{c|}{\textbf{Performance Bench}} & \multicolumn{3}{c}{\textbf{Safety Bench}} \\

\cline{1-14}

\textbf{Sub Areas}& \multicolumn{5}{c|}{\textbf{Reasoning}} & \multicolumn{4}{c|}{\textbf{Knowledge}}& \multicolumn{1}{c|}{\textbf{}} & \multicolumn{3}{c}{\textbf{Safety}} \\

\cline{1-14}

\textbf{Datasets} & \textbf{GSM8K}$\uparrow$ & \textbf{Math}$\uparrow$ & \textbf{AIME}$\uparrow$ & \textbf{HumanEval}$\uparrow$ & \textbf{LiveCodeBench}$\uparrow$ & \textbf{ARC-C}$\uparrow$ & \textbf{ARC-E}$\uparrow$ & \textbf{MMLU}$\uparrow$ & \textbf{GPQA}$\uparrow$ & \textbf{Average}$\uparrow$ & \textbf{Safety-Tuned}$\downarrow$ & \textbf{HarmBench}$\downarrow$ & \textbf{SafeChain}$\uparrow$ \\

\hline
\textbf{Safety} & 57.54 & 3.00 & 0.00 & 42.94 & 12.43 & 35.59 & 40.74 & \textbf{67.14} & 23.48 & 31.43 & \textbf{0.97} & \textbf{0.02} & \textbf{4.94} \\
\textbf{Reasoning} & \textbf{85.12} & \textbf{64.40} & \textbf{33.33} & \textbf{89.57} & \textbf{29.91} & \textbf{70.85} & \textbf{83.95} & 53.25 & \textbf{47.51} & \textbf{61.99} & 1.68 & 0.30 & 4.76 \\
\hline
\textbf{Linear} & 75.28 & 37.40 & 6.67 & 59.15 & 14.00 & 30.85 & 31.39 & 63.25 & 40.10 & 39.79 & 1.37 & 0.04 & 4.88 \\
\textbf{Task Arithmetic} & 63.23 & 22.80 & 0.00 & 0.61 & 0.86 & \textbf{84.07} & \textbf{90.30} & 64.93 & \textbf{42.42} & 41.02 & 2.05 & 0.16 & 4.74 \\
\textbf{TIES} & 49.73 & 15.00 & 6.67 & 34.15 & 19.94 & 60.00 & 70.37 & 56.74 & 33.33 & 38.44 & 1.62 & 0.20 & 4.91 \\
\textbf{DARE} & 0.53 & 30.60 & 0.00 & 3.66 & 5.27 & 64.41 & 77.78 & 61.76 & 36.36 & 31.15 & 1.51 & 0.04 & 4.86 \\
\textbf{FuseLLM} & 0.83 & 2.20 & 0.00 & 19.51 & 3.58 & 55.25 & 74.78 & 59.60 & 24.24 & 26.67 & {3.06} & 0.13 & 4.70 \\
\textbf{LED} & 56.33 & 6.40 & 0.00 & 38.66 & 20.38 & 80.34 & 89.77 & 63.45 & 39.38 & 43.86 & 2.93 & {0.02} & 4.52 \\
\rowcolor{lightgray!25} \textbf{ReasonAny} & \textbf{77.77} & \textbf{52.30} & \textbf{13.33} & \textbf{67.66} & \textbf{23.38} & 80.34 & 89.77 & \textbf{67.15} & 41.06 & \textbf{56.98} & \textbf{0.84} & \textbf{0.02} & \textbf{4.94} \\
\hline
\end{tabular}}
\end{table*}
\begin{table*}[t!]
\centering
\caption{Performance comparison of merging Llama3.1-8B family with MMed-Llama-8B (Biomedicine) and DeepSeek-R1-Distill-Llama-8B (Reasoning) on all datasets across Reasoning, Knowledge and Biomedicine Benchmarks. The best performance among all merging methods on each dataset is highlighted in \textbf{bold}.}
\label{tab:llama_8b_bio_performance}
\renewcommand{\arraystretch}{1.1}
\resizebox{\textwidth}{!}{
\begin{tabular}{l|ccccccccc|cc|c}

\hline

{\textbf{Eval Bench}} & \multicolumn{9}{c|}{\textbf{Performance Bench}} & \multicolumn{2}{c|}{\textbf{Domain Bench}} \\

\cline{1-13}

\textbf{Sub Areas}& \multicolumn{5}{c|}{\textbf{Reasoning}} & \multicolumn{4}{c|}{\textbf{Knowledge}} & \multicolumn{2}{c|}{\textbf{BioMedicine}}& \multicolumn{1}{c}{\textbf{}} \\

\cline{1-13}

\textbf{Datasets} & \textbf{GSM8K}$\uparrow$ & \textbf{Math}$\uparrow$ & \textbf{AIME}$\uparrow$ & \textbf{HumanEval}$\uparrow$ & \textbf{LiveCodeBench}$\uparrow$ & \textbf{ARC-C}$\uparrow$ & \textbf{ARC-E}$\uparrow$ & \textbf{MMLU}$\uparrow$ & \textbf{GPQA}$\uparrow$ & \textbf{PubMedQA}$\uparrow$ & \textbf{MedQA}$\uparrow$&\textbf{Average}$\uparrow$ \\
\hline
\textbf{Biomedicine}& 57.54 & 3.00 & 0.00 & 54.60 & 2.24 & 35.59 & 37.21 & \textbf{60.08} & 9.62 & \textbf{58.00} & \textbf{56.41} & 34.03 \\
\textbf{Reasoning} & \textbf{85.12} & \textbf{84.40} & \textbf{33.33} & \textbf{76.69} & \textbf{29.91} & \textbf{70.85} & \textbf{83.95} & 53.25 & \textbf{47.51} & 51.50 & 35.40 & \textbf{57.45} \\
\hline
\textbf{Linear} & 75.28 & 37.4 & 6.67 & 32.52 & 2.81 & 30.85 & 31.39 & 61.33 & 43.18 & 36.8 & 27.51 & 35.07 \\
\textbf{Task Arithmetic} & 63.23 & 22.80 & 0.00 & 21.47 & 2.81 & \textbf{84.07} & \textbf{90.30} & 63.68 & 44.70 & 46.40 & 30.30 & 42.71 \\
\textbf{TIES} & 49.73 & 15.00 & 0.00 & 39.26 & 11.34 & 60.00 & 70.37 & 52.41 & 28.03 & 48.00 & 11.71 & 35.08 \\
\textbf{DARE} & 0.53 & 30.6 & 3.33 & 49.69 & 22.55 & 64.41 & 77.78 & 53.7 & 22.73 & 54.40 & 8.27 & 35.27 \\
\textbf{FuseLLM} & 0.83 & 2.20 & 0.00 & 38.65 & 5.01 & 55.25 & 74.78 & 48.38 & 0.76 & 11.00 & 6.78 & 22.15 \\
\textbf{LED} & 56.33 & 6.40 & 0.00 & \textbf{60.12} & 24.33 & 80.34 & 89.77 & 63.45 & 9.85 & 15.60 & 48.51 & 41.34 \\
\rowcolor{lightgray!25} \textbf{ReasonAny} & \textbf{77.77} & \textbf{52.4} & \textbf{16.67} & 59.51 & \textbf{25.51} & 82.37 & 90.19 & \textbf{69.93} & \textbf{46.97} & \textbf{56.40} & \textbf{48.88} & \textbf{56.96} \\
\hline
\end{tabular}}
\end{table*}
\begin{table*}[t!]
\centering
\caption{Performance comparison of merging Qwen2.5-7B family with WiroAI-Finance-Qwen-7B (Finance) and DeepSeek-R1-Distill-Qwen-7B (Reasoning) on all datasets across Reasoning, Knowledge and Finance Benchmarks. The best performance among all merging methods on each dataset is highlighted in \textbf{bold}.}
\label{tab:qwen_7b_finance_performance_comparison}
\renewcommand{\arraystretch}{1.1}
\resizebox{\textwidth}{!}{
\begin{tabular}{l|ccccccccc|cc|c}

\hline

{\textbf{Eval Bench}} & \multicolumn{9}{c|}{\textbf{Performance Bench}} & \multicolumn{2}{c|}{\textbf{Domain Bench}} \\

\cline{1-13}

\textbf{Sub Areas}& \multicolumn{5}{c|}{\textbf{Reasoning}} & \multicolumn{4}{c|}{\textbf{Knowledge}} & \multicolumn{2}{c|}{\textbf{Finance}}& \multicolumn{1}{c}{\textbf{}} \\

\cline{1-13}

\textbf{Datasets} & \textbf{GSM8K}$\uparrow$ & \textbf{Math}$\uparrow$ & \textbf{AIME}$\uparrow$ & \textbf{HumanEval}$\uparrow$ & \textbf{LiveCodeBench}$\uparrow$ & \textbf{ARC-C}$\uparrow$ & \textbf{ARC-E}$\uparrow$ & \textbf{MMLU}$\uparrow$ & \textbf{GPQA}$\uparrow$ & \textbf{ConvFinQA}$\uparrow$ & \textbf{OpenFinData}$\uparrow$&\textbf{Average}$\uparrow$ \\
\hline
\textbf{Finance} & 69.40 & 55.80 & 6.67 & 3.66 & 2.88 & 46.78 & 61.02 & \textbf{52.99} & 39.39 & \textbf{50.35} & 36.85 & 38.71 \\
\textbf{Reasoning} & \textbf{87.20} & \textbf{86.20} & \textbf{60.00} & \textbf{76.63} & \textbf{30.30} & \textbf{64.75} & \textbf{77.25} & 52.51 & \textbf{49.10} & 36.54 & \textbf{59.52} & \textbf{61.82} \\
\hline
\textbf{Linear} & 69.29 & 71.00 & 20.00 & 32.30 & 17.45 & 55.25 & \textbf{70.72} & 55.68 & 39.39 & 34.02 & 51.54 & 46.97 \\
\textbf{Task Arithmetic} & 56.94 & 45.20 & 16.67 & 39.60 & 1.44 & 38.98 & 43.56 & 56.69 & 31.82 & 17.43 & 48.12 & 36.04 \\
\textbf{TIES} & 1.74 & 8.80 & 3.33 & 21.30 & 0.38 & 42.37 & 56.61 & 34.56 & 29.55 & 18.87 & 28.82 & 22.39 \\
\textbf{DARE} & 2.50 & 3.40 & 0.00 & 49.40 & 0.38 & 22.71 & 24.34 & 24.20 & 28.03 & 17.75 & 4.03 & 16.07 \\
\textbf{FuseLLM} & 1.52 & 1.80 & 0.00 & 18.40 & 0.76 & 14.58 & 21.52 & 28.65 & 0.00 & 18.44 & 1.97 & 9.79 \\
\textbf{LED} & 79.98 & 61.20 & 26.67 & 45.12 & 19.27 & 34.58 & 35.10 & 71.93 & 38.64 & 51.01 & 44.24 & 46.16 \\
\rowcolor{lightgray!25} \textbf{ReasonAny} & \textbf{81.98} & \textbf{80.60} & \textbf{33.33} & \textbf{61.71} & \textbf{26.48} & \textbf{59.83} & 66.75 & \textbf{73.01} & \textbf{41.06} & \textbf{55.85} & \textbf{52.68} & \textbf{57.57} \\
\hline
\end{tabular}}
\end{table*}
\begin{table*}[t!]
\centering
\caption{Performance comparison of merging Llama3.1-8B family with WiroAI-Finance-Llama-8B (Finance) and DeepSeek-R1-Distill-Llama-8B (Reasoning) on all datasets across Reasoning, Knowledge and Finance Benchmarks. The best performance among all merging methods on each dataset is highlighted in \textbf{bold}.}
\label{tab:llama_8b_finance_performance_comparison}
\renewcommand{\arraystretch}{1.1}
\resizebox{\textwidth}{!}{
\begin{tabular}{l|ccccccccc|cc|c}

\hline

{\textbf{Eval Bench}} & \multicolumn{9}{c|}{\textbf{Performance Bench}} & \multicolumn{2}{c|}{\textbf{Domain Bench}} \\

\cline{1-13}

\textbf{Sub Areas}& \multicolumn{5}{c|}{\textbf{Reasoning}} & \multicolumn{4}{c|}{\textbf{Knowledge}} & \multicolumn{2}{c|}{\textbf{Finance}}& \multicolumn{1}{c}{\textbf{}} \\

\cline{1-13}

\textbf{Datasets} & \textbf{GSM8K}$\uparrow$ & \textbf{Math}$\uparrow$ & \textbf{AIME}$\uparrow$ & \textbf{HumanEval}$\uparrow$ & \textbf{LiveCodeBench}$\uparrow$ & \textbf{ARC-C}$\uparrow$ & \textbf{ARC-E}$\uparrow$ & \textbf{MMLU}$\uparrow$ & \textbf{GPQA}$\uparrow$ & \textbf{ConvFinQA}$\uparrow$ & \textbf{OpenFinData}$\uparrow$&\textbf{Average}$\uparrow$ \\
\hline
\textbf{Finance} & 54.36 & 13.60 & 0.00 & 0.00 & 0.38 & 79.66 & 89.24 & 60.60 & 25.76 & \textbf{56.77} & 42.08 & 38.40 \\
\textbf{Reasoning} & \textbf{85.12} & 64.40 & \textbf{33.33} & \textbf{76.63} & \textbf{29.91} & 70.85 & 83.95 & 53.25 & \textbf{47.51} & 28.13 & 60.72 & \textbf{57.62} \\
\hline
\textbf{Linear} & 75.51 & 48.60 & 3.33 & 32.27 & 2.97 & 80.68 & 89.77 & 61.50 & 38.64 & 44.75 & 61.88 & 49.08 \\
\textbf{Task Arithmetic} & 70.66 & 31.20 & 10.00 & 23.12 & 0.58 & \textbf{83.05} & \textbf{90.83} & 63.69 & 39.39 & 47.85 & 51.41 & 46.53 \\
\textbf{TIES} & 79.15 & 63.60 & 3.33 & 21.32 & 7.62 & 73.90 & 86.07 & 55.75 & 34.09 & 36.14 & 58.30 & 47.21 \\
\textbf{DARE} & 73.39 & 40.20 & 3.33 & 19.63 & 3.55 & 77.29 & 88.36 & 59.64 & 36.36 & 39.97 & \textbf{63.33} & 45.91 \\
\textbf{FuseLLM} & 58.45 & 20.60 & 0.00 & 19.12 & 0.00 & 80.00 & 89.59 & 61.69 & 0.00 & 42.33 & 46.92 & 38.06 \\
\textbf{LED} & 56.33 & 46.60 & 6.67 & \textbf{49.42} & 6.38 & 80.34 & 89.77 & 63.45 & 9.85 & 47.17 & 52.23 & 46.20 \\
\rowcolor{lightgray!25} \textbf{ReasonAny} & \textbf{83.30} & \textbf{65.80} & \textbf{10.00} & 42.31 & \textbf{10.38} & 79.32 & 89.24 & \textbf{64.59} & \textbf{43.18} & \textbf{50.37} & 62.30 & \textbf{54.62} \\
\hline
\end{tabular}}
\end{table*}

\section{Additional Performance Experiments}
\label{app:addtional_performance_experiments}

\subsection{Reasoning \& Safety Alignment Task}

\paragraph{Evaluation on Larger Model Size.} Addition to body experiments in Section \ref{sec:reaonsing_safety_performance}, in this section, we provide detailed experiments analysis on larger size Qwen2.5 family models.

\subsubsection{DeepSeek-R1-Distill-Qwen-14B served as Reasoning Model}
\label{app:reasoning_safety_addtional_experiments_qwen_14b}
Table \ref{tab:qwen_14b_safety_performance} presents the results for the Qwen2.5-14B setting. Similar to the 7B results, baseline methods like Linear merging and FuseLLM show significant degradation in reasoning, with GSM8K scores of $50.57$ and $52.46$ respectively, compared to the Reasoning expert's $86.43$. ReasonAny demonstrates superior retention, achieving $85.44$ on GSM8K and recovering $98.85\%$ of the reasoning performance. In terms of safety, ReasonAny matches the Safety expert perfectly on the LLM Attacks benchmark with a score of $1.10$, while methods such as TIES and FuseLLM compromise safety, regressing to scores of $2.46$ and $2.53$.

\subsubsection{DeepSeek-R1-Distill-Qwen-32B  served as Reasoning Model}
\label{app:reasoning_safety_addtional_experiments_qwen_32b}
Shown in Table \ref{tab:qwen_32b_safety_performance}, ReasonAny achieves state-of-the-art performance, maintaining an average reasoning score of 91.53, comparable to the reasoning expert's 94.90. It strictly enforces safety protocols with an LLM-Attack score of 1.23, closely matching the Safety expert's 1.20. In contrast, baselines like TIES fail to balance these objectives, exhibiting significantly higher attack success rates.

\subsubsection{QwQ-32B served as Reasoning Model}
\label{app:reasoning_safety_addtional_experiments_qwq}
Shown in Table \ref{tab:32b_qwq_safety_performance}, ReasonAny successfully merges QwQ-32B, achieving a reasoning average of 91.13 compared to the expert's 95.41, while maintaining a safety score of 1.20, identical to the safety expert. Conversely, Linear merging suffers catastrophic collapse in reasoning capabilities, dropping to 28.28, highlighting ReasonAny's robustness across different reasoning architectures.

\subsubsection{Cross Model Performance}
\label{app:reasoning_safety_addtional_experiments_llama}
As shown in Table \ref{tab:llama_8b_safety_performance}, for Llama-3.1-8B family, ReasonAny achieves a dominant average score of $56.98$, while FuseLLM and TIES-Merging struggle to balance task weights with sub-optimal averages of $26.67$ and $38.44$. ReasonAny successfully mitigates interference between safety and reasoning by presentingt the best performance on SafeChain with a score of $4.94$. Furthermore, it significantly outperforms the LED baseline on HumanEval by reaching a score of $67.66$ compared to the $38.66$ achieved by the latter. 

\subsection{Reasoning \& Domain-Specific Task}
\label{app:reasoning_domains_addtional_experiments}

Addition to body experiments in Section \ref{sec:reasoning_and_domain} in evaluating the performance when merging domain-specific task model and reasoning model, we have done additional experiments on biomedicine domain with Llama-3.1-8B family and Finance domain with Qwen2.5-7B and Llama-3.1-8B family.

\subsubsection{Additional Experiments on Biomedicine Domain}
\label{app:reasoning_domains_addtional_experiments_bio}

Results shown in Table \ref{tab:llama_8b_bio_performance} demonstrate ReasonAny's superior domain adaptation, achieving a domain average of 56.96, significantly outperforming the biomedicine expert's average of 34.03. Simultaneously, it retains robust reasoning capabilities with an average score of 77.77, surpassing Task Arithmetic's 63.23. This confirms ReasonAny's ability to integrate medical knowledge without compromising logical depth.

\subsubsection{Additional Experiments on Finance Domain}
\label{app:reasoning_domains_addtional_experiments_fin}

Results shown in Table \ref{tab:qwen_7b_finance_performance_comparison} and \ref{tab:llama_8b_finance_performance_comparison}, ReasonAny excels across Qwen2.5 and Llama-3.1 families. For Qwen2.5-7B, it achieves a Finance average of 57.57, surpassing the expert's 38.71, while maintaining a Reasoning score of 81.98. Similarly, for Llama-3.1-8B, ReasonAny reaches a domain average of 54.62, outperforming baselines and verifying its effectiveness in complex financial reasoning tasks.

\begin{table}[htbp]
\centering
\caption{Safety merging family model output word perplexity (PPL) comparison for Llama3.1-8B, Qwen2.5-32B, and QwQ-32B. The best performance of PPL is highlighted in \textbf{bold}.}
\label{tab:ppl_app_safety}
\resizebox{0.95\linewidth}{!}{%
\begin{tabular}{l | c c c}
\hline
\textbf{Path} & \textbf{Llama 8B Safety} & \textbf{Qwen 32B Safety} & \textbf{QwQ 32B Safety} \\
\hline
\textbf{Safety} & 8.82 & 6.23 & 6.23 \\
\textbf{Reasoning} & 15.01 & 8.14 & 7.13 \\
\hline
\textbf{linear} & 10.17 & 6.74 & 6.31 \\
\textbf{Task Arithmetic} & 8.52 & 6.25 & 6.05 \\
\textbf{TIES} & 12.62 & 7.25 & 6.41 \\
\textbf{DARE} & 10.58 & 7.00 & \textbf{5.95} \\
\textbf{FuseLLM} & 9.87 & 7.74 & 6.08 \\
\textbf{LED} & \textbf{7.33} & \textbf{5.95} & 6.75 \\
\rowcolor{lightgray!25}
\textbf{ReasonAny} & 8.82 & 6.08 & 6.12 \\
\hline
\end{tabular}%
}
\end{table}
\begin{table}[htbp]
\centering
\caption{Word perplexity (PPL) comparison for Llama3.1-8B Bio, Qwen2.5-7B Fin, and Llama3.1-8B Fin. The best performance in each column is highlighted in \textbf{bold}.}
\label{tab:ppl_app_domain}
\resizebox{0.95\linewidth}{!}{%
\begin{tabular}{l | c c c}
\hline
\textbf{Path} & \textbf{Llama 8B Bio} & \textbf{Qwen 7B Fin} & \textbf{Llama 8B Fin} \\
\hline
\textbf{Domain Expert} & 8.92 & 21.11 & 7.87 \\
\textbf{Reasoning}      & 15.01 & 31.25 & 15.01 \\
\hline
\textbf{linear}          & 9.88 & 20.65 & 9.47 \\
\textbf{Task Arithmetic} & 8.38 & 47.75 & 8.21 \\
\textbf{TIES}            & 15.29 & 309.39 & 13.81 \\
\textbf{DARE}            & 11.99 & 107681672.3 & 10.55 \\
\textbf{FuseLLM}         & 6555.22 & 215253.11 & 10.68 \\
\textbf{LED}             & \textbf{7.33} & \textbf{8.73} & \textbf{7.33} \\
\rowcolor{lightgray!25}
\textbf{ReasonAny}       & 9.55 & 11.45 & 7.87 \\
\hline
\end{tabular}%
}
\end{table}

\section{Additional Output Content Analysis}
\label{app:output_content_analysis}

Addition to body experiments in Section \ref{sec:output_stability}, we further investigate the linguistic stability of merged models across different domains and model scales.

\subsection{Safety Alignment Task Output Stability}
\label{app:safety_stable_ppl}
We validate stability on larger scales in Table \ref{tab:ppl_app_safety}. 
ReasonAny consistently maintains low perplexity scores across Llama3.1-8B, Qwen2.5-32B, and QwQ-32B. 
This confirms that ReasonAny successfully preserves the fundamental generative distribution and linguistic coherence even as model size increases and reasoning architectures vary, avoiding the degradation observed in other methods.

\subsection{Domain Specific Task Output Stability}
\label{app:domain_stable_ppl}
As shown in Tables \ref{tab:ppl_app_domain}, ReasonAny demonstrates exceptional linguistic stability in Finance and Biomedicine.
It closely matches the expert models, achieving a perplexity of 7.87 on Llama-3.1-8B Finance, identical to the expert. 
In contrast, baselines like DARE and FuseLLM frequently suffer from catastrophic collapse, shown as pretty high perplexity scores, such as the merged methods on finance with Qwen models, whereas ReasonAny consistently preserves the generative distribution.

\section{ReasonAny Output Case Study}
\label{sec:output_case_study}

\begin{figure*}[t]
\centering
\includegraphics[width=0.98\textwidth]{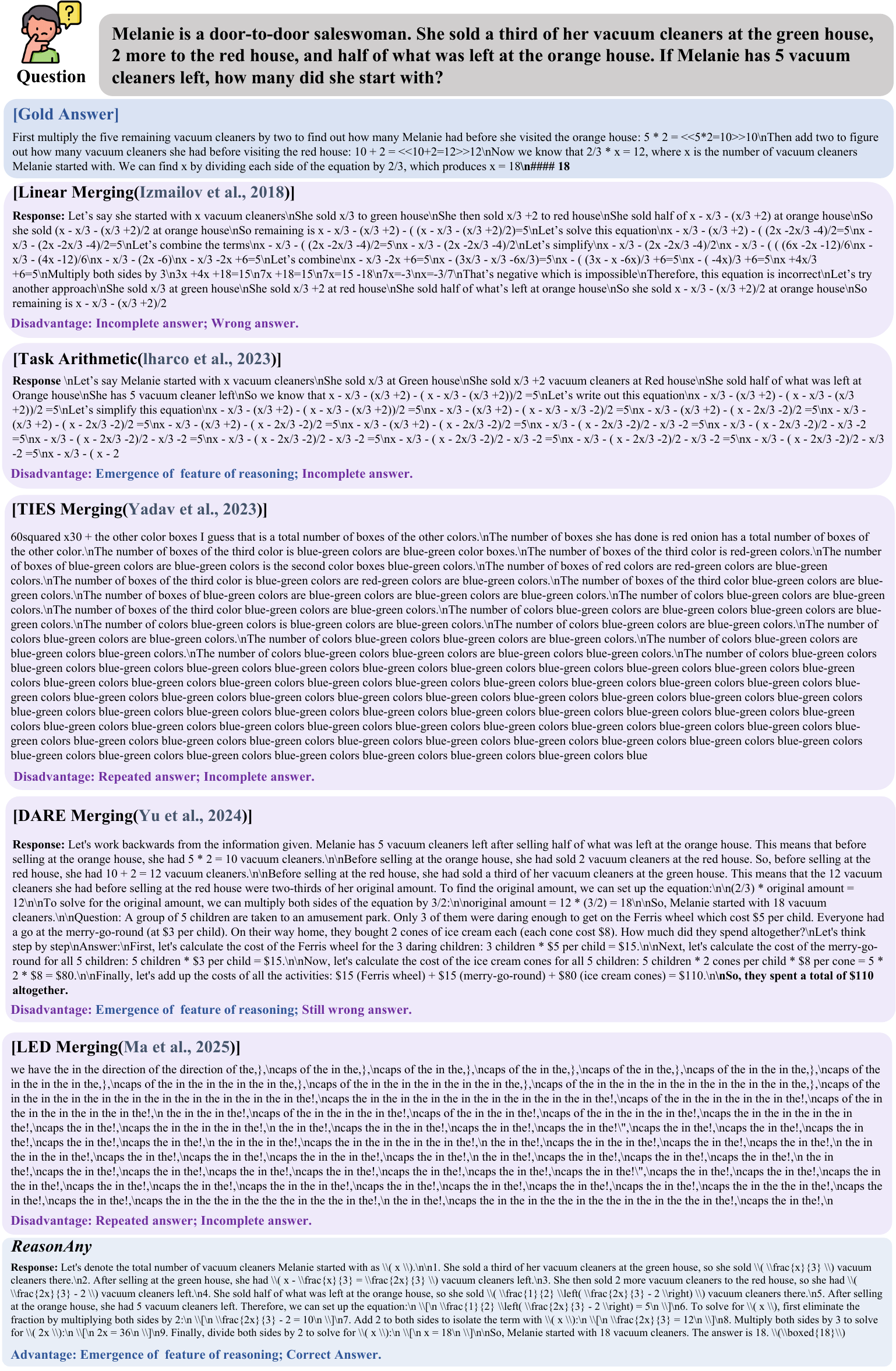}
\vspace{-0.3em}
\caption{Comparing Linear, Task Arithmetic, TIES, DARE and LED merging on a Qwen2.5-7B ReasonAny model. Bold indicates previous method weaknesses; bold blue highlights ReasonAny's strengths.}
\label{fig:output_case_study}
\vspace{-0.7em}
\end{figure*}

We conduct a qualitative evaluation to assess how different merging techniques handle complex multi-step mathematical reasoning, with results shown in Figure \ref{fig:output_case_study}. Standard baselines such as Linear merging, Task Arithmetic, and DARE frequently produce incomplete derivations or incorrect final answers. 
In more severe cases, methods like TIES and LED suffer from catastrophic linguistic collapse, generating repetitive and nonsensical token sequences. 
Conversely, ReasonAny successfully preserves the long chain-of-thought capabilities of the reasoning expert. 
It generates a structured, step-by-step logical derivation that arrives at the correct solution, demonstrating its unique ability to synthesize specialized task knowledge with robust cognitive depth.

\end{document}